\documentclass[11pt]{article}

\usepackage[final]{acl}

\usepackage{times}
\usepackage{latexsym}
\usepackage[T1]{fontenc}
\usepackage[utf8]{inputenc}
\usepackage{microtype}
\usepackage{inconsolata}
\usepackage{booktabs}
\usepackage{graphicx}
\usepackage{multirow}
\usepackage{amsmath}
\usepackage{amssymb}
\usepackage{xcolor}
\usepackage{capt-of}
\usepackage{enumitem}
\usepackage[skins,breakable]{tcolorbox}

\definecolor{darkblue}{rgb}{0, 0, 0.5}
\definecolor{promptcolor}{HTML}{1e40af}
\definecolor{promptbg}{HTML}{eff6ff}
\definecolor{responsebg}{HTML}{f0fdf4}
\definecolor{responsecolor}{HTML}{166534}

\newtcolorbox[auto
counter,number within=section]{prompt}[1][]{
    colbacktitle=promptcolor,
    colback=promptbg,
    colframe=promptcolor,
    fonttitle=\bfseries\small,
    fontupper=\footnotesize,
    boxsep=5pt,
    left=3pt,
    right=3pt,
    top=2pt,
    bottom=2pt,
    boxrule=0.8pt,
    enhanced,
    breakable,
    skin first=enhanced,
    skin middle=enhanced,
    skin last=enhanced,
    #1,
}

\newtcolorbox[auto counter,number within=section]{response}[1][]{
    colbacktitle=responsecolor,
    colback=responsebg,
    colframe=responsecolor,
    fonttitle=\bfseries\small,
    fontupper=\footnotesize,
    boxsep=5pt,
    left=3pt,
    right=3pt,
    top=2pt,
    bottom=2pt,
    boxrule=0.8pt,
    enhanced,
    breakable,
    skin first=enhanced,
    skin middle=enhanced,
    skin last=enhanced,
    #1,
}
\hypersetup{colorlinks=true, citecolor=darkblue, linkcolor=darkblue, urlcolor=darkblue}

\newcommand{\mqud}{\textsc{MQUD}}

\title{Multimodal QUD: \\ Inquisitive Questions from Scientific Figures}

\author{\textbf{Yating Wu}$^{1}$, \textbf{William Rudman}$^{1}$, \textbf{Venkata S.\ Govindarajan}$^{2}$\\
\textbf{Alexandros G.\ Dimakis}$^{3}$, \textbf{Junyi Jessy Li}$^{1}$\\[0.4em]
\normalfont
$^{1}$The University of Texas at Austin \quad
$^{2}$Ithaca College \quad
$^{3}$UC Berkeley, BespokeLabs.ai\\
\normalfont\texttt{yating.wu@utexas.edu}}

\begin{document}
\maketitle

\begin{abstract}
Discourse comprehension in complex documents often involves continuously posing and resolving Questions Under Discussion (QUDs). While QUD frameworks have so far focused on text, scientific literature is inherently multimodal: figures convey discourse goals distinct from their textual counterparts, thus invoking implicit questions that the surrounding text answers. In scientific discovery, knowing the right questions to ask is as important as knowing how to answer them, yet this capability remains largely absent in current models. In this work, we extend QUD to multimodal discourse in scientific literature, targeting questions evoked by figures that are (1) \emph{inquisitive}, i.e., not resolved in the prior context; (2) \emph{salient}, i.e., relevant to the paper's research claims and addressed later in the paper; (3) grounded in visual insights. To benchmark model capability to generate such questions, we introduce \mqud{}, a dataset of 1,250 figure-evoked questions from 56 scientific papers, including 708 questions annotated by the original authors of these papers. Our experiments show that open-source VLMs such as Qwen 3.5 predominantly ask questions that can be answered by the figure alone. Fine-tuning on \mqud{} teaches the model to ask scientifically relevant, inquisitive questions that target the role a figure plays in the paper's argument.
\end{abstract}

\section{Introduction}
\label{sec:intro}

    The act of inquiry is a signature of human curiosity \cite{loewenstein1994curiosity}. It is also highly relevant to  discourse processing: in theories of Questions Under Discussion, discourse is organized around implicit questions that each claim raises or partly answers \citep{vankuppevelt1995discourse,roberts2012information}.
In expectation-driven accounts of QUD \cite{kehler2017expectation}, discourse is forward-looking, where new information evokes (implicit) potential questions whose answers the readers anticipate \cite{westera2020tedq, ko-etal-2020-inquisitive}. Indeed, writers tend to answer questions deemed salient by the readers even when no explicit communication exists \cite{wu2025qsalience}. From this point of view, these potential questions are highly valuable: they directly capture curiosity, tend to be high-level \cite{dcqa},
and answering them would advance the overall comprehension of the discourse at hand. 

Even without an existing answer, such questions can have profound implications in creative applications (e.g., scientific discovery), where knowing the right questions to ask is as important as knowing how to answer them \cite{brainard2025artificial}. Despite the popularity of AI science agents \citep{lu2024aiscientist}, such capabilities remain largely absent in models \citep{wang2024scibench}. In science, the questions worth asking often arise from figures: scientific literature is inherently multimodal (as research papers typically include diagrams, figures, tables, etc), whereas existing work on QUD, both theoretical and empirical, has so far been exclusively for text. Existing multimodal benchmarks primarily test chart understanding \citep{chartmuseum, masry2022chartqa, roberts2024scifibench}, focus on retrieval-based questions \citep{pramanick2024spiqa} and do not evaluate a question's depth or salience \citep{multimodal_arxiv}.

\begin{figure*}[t]
\centering
\includegraphics[width=\linewidth]{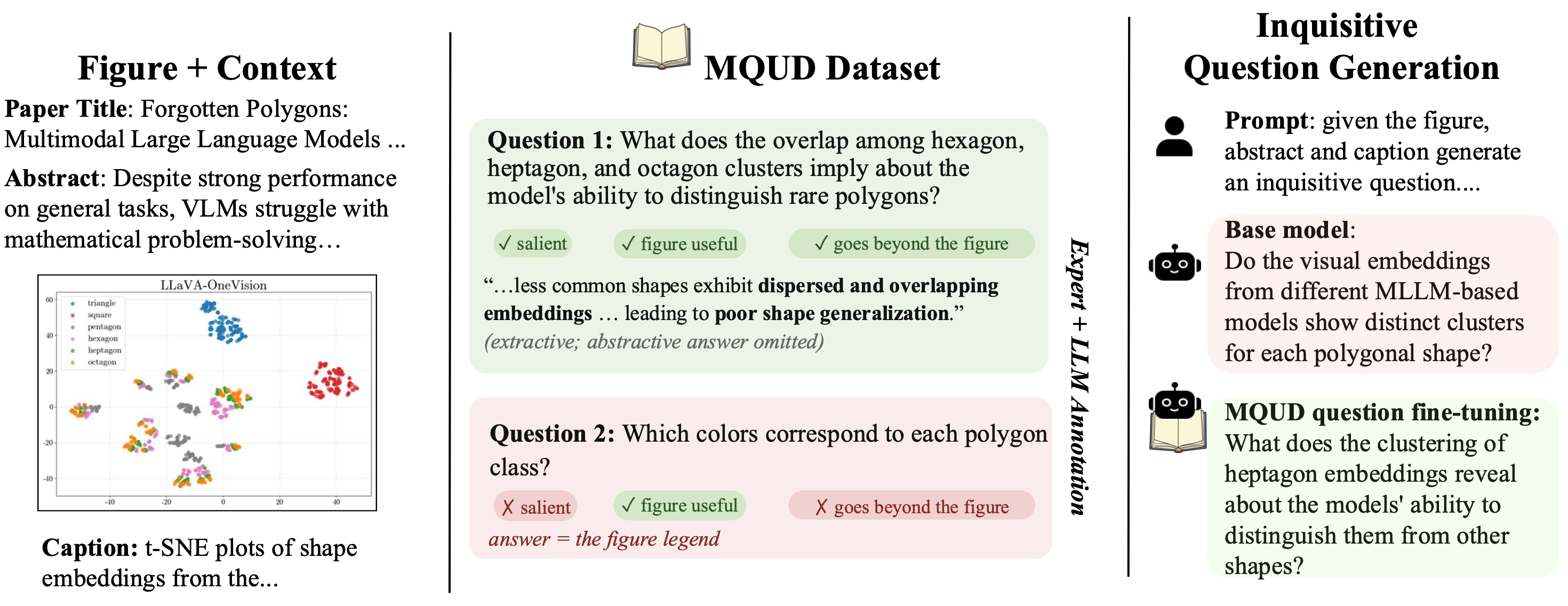}
\caption{This work extends QUD to multimodal scientific discourse. Given paper title, abstract, a figure and its caption, the MQUD dataset includes gold (expert) and silver (LLM-as-judge) labels for expectation-driven QUDs, across seven dimensions: Question~1 is salient and goes beyond the figure; Question~2 is answerable from the figure alone. The right panel contrasts generation before and after fine-tuning on \mqud{}.}
\label{fig:concept}
\end{figure*}

This paper extends the QUD framework to the multimodal setting with a novel dataset, which we show to be useful for generating inquisitive and insightful questions that arise from figures in scientific research papers.
Generating such questions
from figures poses two challenges: (1) sourcing questions that go beyond visual extraction and (2) evaluating their scientific value.
We focus on figures as they serve critical communicative functions in research papers: they illustrate a paper's key motivation, provide nuanced interpretations of data, and show findings in ways that text alone is unable to communicate \citep{larkin1987diagram,lee2018viziometrics}. Figures are prominent question triggers: When readers see accuracy curves diverging, they may ask \emph{why}; when they see clusters in an embedding space, they may ask \emph{what is the significance of these clusters}.
Many of these questions
are evoked
because the text sets expectations that the figure challenges, or because the figure reveals a pattern that the text has not yet explained.
Answering these questions often requires a deep understanding of the papers' research goal.

\begin{table*}[t]
\centering
\vspace{0.2\baselineskip}
\begingroup
\small
\setlength{\tabcolsep}{4pt}
\renewcommand{\arraystretch}{1.12}
\begin{tabular}{@{}l cccccc@{}}
\toprule
& \textbf{ChartQA} & \textbf{SciGraphQA} & \textbf{SPIQA} & \textbf{SciVQA} & \textbf{VeriSciQA} & \textbf{\mqud{}} \\
\midrule
Question source & Chart content & Paper context & Fig.+citing text & Fig.+caption & Citing text & Fig--text interaction \\
Answer type & Short values & Free-form & Free-form & Short answers & Multiple choice & Text-grounded expl. \\
Answer source & Chart alone & Chart+caption & Fig.+paper text & Fig.+caption & Figure & Paper text \\
Fig.\ specificity & Not tested & Not tested & Not tested & Not tested & Not tested & Verified \\
Annotation & Correctness & -- & Correctness & Semi-automatic & Auto-verified & 7 dimensions \\
\bottomrule
\end{tabular}
\endgroup
\caption{Comparison of scientific figure QA datasets. \mqud{} is the first to target questions arising from figure--text interaction, with verified figure specificity and multi-dimensional annotation. Additional depth analysis in Appendix~\ref{app:depth-analysis}.}
\label{tab:dataset-comparison}
\end{table*}

First, we formalize our framework for multimodal QUDs in \S\ref{sec:framework} and build \mqud{} (Figure~\ref{fig:concept}, middle panel), a dataset of 1,250 multimodal questions from 56 scientific papers across three domains: NLP, Machine Learning, and astronomy (\S\ref{sec:dataset}). Since scientific utility can be subjective, 
a subset of 708 questions
contain salience and quality judgments \textbf{from the original authors of these papers}.
Our annotations show that QUD type is systematically associated with a figure's discourse role (\S\ref{sec:annotation-analysis}). 

In \S~\ref{sec:experiments} we use \mqud{} to develop models that generate high-quality inquisitive questions from figures (Figure~\ref{fig:concept}, right panel). Concretely, we find that fine-tuning Qwen 3.5 (a VLM) on \mqud{} increases the rate of salient questions by ~6\% and produces questions that cannot be answered by the figure alone by ~8\%. By moving beyond existing work for generating questions from scientific figures, which keeps only questions the figure itself can answer, our inquisitive question generation framework can facilitate scientific discovery by posing research questions yet to be explored. 

\section{Related Work}
\label{sec:related}

\paragraph{Question Under Discussion and theoretical foundations of this work.}
QUD theory models discourse as implicit questions raised and resolved by assertions \citep{vankuppevelt1995discourse,roberts2012information}. Recent work operationalizes this through structured prediction \citep{ko2022discourse,dcqa} and evaluation benchmarks \citep{wu2023qudeval}.

The relevance between QUD and inquisitive, reader-generated questions is supported by psycholinguistic experiments of expectation-driven QUD. \citet{kehler2017expectation} show that discourse processing is \textbf{proactive} where readers raise (implicit) questions whose types are influenced by prior context (their experiment specifically looked at causality). Empirical work by \citet{wu2025qsalience} find that demographically homogeneous reader groups have a high agreement regarding the salience of inquisitive questions given the context, and \emph{without} seeing upcoming discourse. Furthermore, questions rated high for salience are more likely to be answered later in the same document by the authors, despite an utter lack of communication channels.

These lines of work provide the foundation of our conceptualization: QUD is a suitable framework for eliciting inquisitive questions from context; and while many \emph{valid} questions can be generated, the QUD theory is particularly suited for eliciting salient questions. That said, existing work focuses on text, and existing data is only from news texts or TED talks \citep{dcqa,westera2020tedq}. A key innovation in this work is to extend the QUD theory to multimodal, where the common ground consists mainly of figures and visualizations; and extend the notion of salience to scientific utility in research papers.

\paragraph{Scientific figure understanding.}
Benchmarks for scientific figure QA, including ChartQA \citep{masry2022chartqa}, FigureQA \citep{kahou2018figureqa}, PlotQA \citep{methani2020plotqa}, and SciFIBench \citep{roberts2024scifibench}, focus on perceptual questions answerable from the figure alone. Text-only scientific QA datasets such as QASA \citep{lee2023qasa} and PeerQA \citep{baumgartner2025peerqa} focus on document-level question answering from paper text alone. \mqud{} targets this gap; as Table~\ref{tab:dataset-comparison} shows, it is the only dataset whose questions arise from the interaction between figure and text.

\paragraph{Generating Questions for Scientific Figures} While several works have focused on generating questions for scientific visuals sourced from research papers, they differ from MQUD in several key ways. \citet{sci-vqa} focus on questions that are answerable from only the figure and caption.  \citet{sci-graph-qa} condition generation on the title, abstract, and the paragraph referencing the figure, but only retain questions that contain a set of keywords related to figure components, such as ``figure'' or ``plot''. In VeriSciQA, questions and answers are generated purely from figure-citing paragraphs and then verified using the referenced figures \cite{veri-sci-qa}. After filtering, VeriSciQA only retains questions whose answer is present in the figure, removing the figure-triggered, text-resolved questions we target. \citet{pramanick2024spiqa} generate questions from a figure together with the passages that reference it, which is the closest setup to ours. Their questions are filtered by the dataset authors for whether they are too trivial or too specific, but not for whether they matter to the paper's research claims. Beyond generated questions, cPAPERS \citep{sundar2024cpapers} collects questions from OpenReview rebuttals that reference figures, which seek clarification about the figure itself. MISS-QA \citep{zhao2025missqa} provides expert-written questions on schematic diagrams whose answers require the paper text, but the questions are written to test comprehension. Importantly, none of the current datasets target the implicit questions figures \emph{raise} in discourse. 

\section{Multimodal QUD Framework}
\label{sec:framework}

In our framework (Figure~\ref{fig:concept}),
a figure can \emph{trigger} a potential question \citep{onea2016potential} that text alone would not raise (e.g., a visual pattern demands explanation, or a comparison becomes apparent only when data is plotted side by side). If the answer is grounded in the paper text, this question becomes a QUD \citep{wu2025qsalience}. By definition, this question is \emph{inquisitive} since it is not answerable by the figure and the prior context.

While some QUD theories, such as \citet{roberts2012information} or \citet{riester2019constructing}, model discourse as a stack or tree of questions, we follow \citet{westera2020tedq}'s expectation-driven approach that treats evoked questions independently. This independence was also assumed in \citet{ko2022discourse}'s dependency structure, though their work generates questions retrospectively from the answers.

\subsection{Structure}
We define a multimodal question, $Q_F$, as one triggered by a figure $F$, its caption and its prior context $C_F$.
We define $C_F$ as the paper title and abstract. We use a minimalist context, since academic reading is frequently non-linear \citep{bazerman1985physicists}, making it difficult to determine what a reader has already encountered when they view a figure.

For $Q_F$ to be a multimodal QUD, \textbf{it must be answered in the paper}. We consider two answer types for $Q_F$: an \emph{extractive} answer, $E_F$, consisting of the paper passages that directly answer $Q_F$, and an \emph{abstractive} answer, $A_F$, that summarizes the content of $E_F$. The abstractive form is needed because the extractive answer, $E_F$, may comprise non-consecutive passages. 

\subsection{Question types}
\label{sec:qud-types}

We reinterpret six question types from \citet{cao2021controllable}'s taxonomy to apply to multimodal discourse. Each question receives exactly one mutually exclusive primary type in the generation schema. The types describe different information gaps between the text and figure $F$ (see Appendix~\ref{app:examples} for examples):

\begin{itemize}[nosep, leftmargin=*]
    \item \textbf{Cause}: \emph{``Why does X occur?''} $F$ shows a pattern; the reader asks for explanation.
    \item \textbf{Comparison}: \emph{``How do X and Y differ?''} $F$ displays elements side by side.
    \item \textbf{Extent}: \emph{``How much?''} $F$ quantifies a phenomenon.
    \item \textbf{Consequence}: \emph{``What happens when?''} $F$ shows an outcome or effect.
    \item \textbf{Procedural}: \emph{``How is X achieved?''} $F$ illustrates a method or pipeline.
    \item \textbf{Concept}: \emph{``What does X represent?''} $F$ uses a visual representation (e.g., clusters, gradients).
\end{itemize}

\citet{cao2021controllable}'s original taxonomy includes nine question types. We exclude three (Verification, Disjunctive, Judgmental) that rarely arise in scientific figure interpretation.\footnote{Applying their pretrained classifier to all 1,250 questions: agreement is 97\% for cause, 93\% for extent, 80\% for consequence; lower for comparison (69\%), concept (71\%), procedural (70\%).} We will show in \S\ref{sec:annotation-analysis} that comparison and extent questions tend to be more likely to be answered by the figure alone. 
Cause, consequence, procedural, and concept questions tend to more frequently have answers located in the text rather than the figures themselves.

\section{The \mqud{} Dataset}
\label{sec:dataset}

MQUD consists of 1,250 multimodal questions ($Q_F$) along with their corresponding answers $(E_F, A_F)$, generated from 243 scientific figures sourced from 56 papers across 3 scientific domains: NLP, Machine Learning, and astronomy. Of these questions, 708 are annotated by the original authors of the papers, with LLM-judge labels used for the remaining 542. Each question-answer pair is annotated along seven dimensions: salience, figure usefulness, figure answerability, answer correctness, answer quality, figure type and grammar (\S\ref{sec:annotation}). Of the 1,250 questions, 453 are verified to be QUDs (225 by the authors, Table~\ref{tab:annotation-dist}; 228 by the judge).
Note that \mqud{} retains multimodal questions that are not QUDs, since these can be used as hard negatives for future model development (e.g., for preference learning). 

\subsection{Candidate QUD Generation}
\label{sec:generation}
To generate candidate QUDs in \mqud{}, we follow prior work \cite{dcqa} and provide models with the answer scope. Here, we scope the answers to be within paragraphs that refer to $F$; we call these paragraphs $P_F$. On average, $P_F$ spans 1.5 paragraphs, 8.7 sentences, and 202 words. Note that this reflects our design choice that the answer scope of these QUDs in \mqud{} will be within $P_F$, although this is already a step above prior work where the answers are scoped to be single sentences. We leave even larger-scoped QUDs for future work.
Specifically,
\[(Q_F, A_F, E_F) \sim \pi_{LLM}(\cdot|F, C_F, P_F)\]
where $E_F\subseteq P_F$.

We use GPT-4o as the LLM here. For each of the 56 papers, we select figures from the main body beginning with the results section, excluding introductory and appendix figures. This focuses \mqud{} on figures that present key results and analysis rather than overview figures, which typically appear earlier in the paper. 
We generate at least five candidate question--answer pairs per figure and keep the ones that pass the filters below. Table~\ref{tab:dataset-stats} summarizes the dataset, and Appendix~\ref{app:prompts} gives the full generation prompt.

\paragraph{Quality filter}
The filter does an automatic pass to remove clear cases where the answers are out of the specified scope or of low quality. We check the returned evidence against $P_F$ and discard the pair when the answer does not come from $P_F$. We also filter for abstractive answers $A_F$ between 20 and 120 words. The lower bound removes fragments and uninformative answers, while the upper bound limits answers that copy the paragraph. 
We removed questions that do not reference figure content, and de-duplicate semantically redundant questions within each figure. Table~\ref{tab:pipeline-roles} summarizes each model's role in the pipeline.
The resulting dataset contains \textbf{1,250} questions across 243 figures (Table~\ref{tab:dataset-stats}).

\begin{table}[t]
\centering
\small
\begin{tabular}{@{}lr@{}}
\toprule
\textbf{Statistic} & \textbf{Value} \\
\midrule
Total questions & 1,250 \\
Papers & 56 \\
Unique figures & 243 \\
Median questions / figure & 5 \\
Mean questions / figure & 5.1 \\
Answer $A_F$ length (words) & 49 \\
Evidence $E_F$ length (words) & 202 \\
\midrule
\textbf{Question type distribution} & \\
\quad Cause & 295 (24\%) \\
\quad Comparison & 233 (19\%) \\
\quad Extent & 227 (18\%) \\
\quad Consequence & 192 (15\%) \\
\quad Concept & 160 (13\%) \\
\quad Procedural & 143 (11\%) \\
\bottomrule
\end{tabular}
\caption{Summary statistics for \mqud{}.}
\label{tab:dataset-stats}
\end{table}

\begin{table}[t]
\centering
\small
\begin{tabular}{@{}llrr@{}}
\toprule
\textbf{Dimension} & \textbf{Value} & \textbf{N} & \textbf{\%} \\
\midrule
\multirow{3}{*}{Fig.\ usefulness} & Useful & 624 & 88.1 \\
& Not useful & 62 & 8.8 \\
& Missing & 22 & 3.1 \\
\midrule
\multirow{3}{*}{Fig.\ answerability} & Yes & 350 & 49.4 \\
& No & 319 & 45.1 \\
& Missing & 39 & 5.5 \\
\midrule
\multirow{3}{*}{Salience} & Salient & 634 & 89.5 \\
& Not salient & 63 & 8.9 \\
& Missing & 11 & 1.6 \\
\midrule
\multirow{3}{*}{Ans.\ correctness} & Accept. & 592 & 83.6 \\
& Not accept. & 106 & 15.0 \\
& Missing & 10 & 1.4 \\
\midrule
\multirow{3}{*}{Multimodal QUD} & Yes & 225 & 31.8 \\
& No & 440 & 62.1 \\
& Missing & 43 & 6.1 \\
\bottomrule
\end{tabular}
\caption{Label distribution of expert annotations over the 708 expert-annotated questions. Questions missing any defining label count as Missing in the QUD row.}
\label{tab:annotation-dist}
\end{table}

\subsection{Expert annotation}
\label{sec:annotation}

We recruited 17 original paper authors as expert annotators across all three domains, and paid \$20 per hour for evaluating candidate questions across seven dimensions:

\begin{enumerate}[nosep, leftmargin=*]
  \item \textbf{Salience}: Is this question relevant to the paper's main argument? Salient or not salient.
  \item \textbf{Figure usefulness}: Is the figure necessary to resolve this question? Useful or not useful.
  \item \textbf{Figure answerability}: Can the figure and caption alone answer the question, without the body text? Yes or no.
  \item \textbf{Answer correctness}: Is the generated answer factually accurate? Acceptable or not acceptable.
  \item \textbf{Answer quality}: Overall quality of the answer. High or low.
  \item \textbf{Figure type}: What kind of information does the figure convey? Result, data, method, comparison, or other.
  \item \textbf{Question grammar}: Linguistic quality of the question. Acceptable or not acceptable.
\end{enumerate}

Additionally, annotators could edit a candidate question in a free-text field to improve grammar or salience before submitting their labels; they did so for 80 questions.

Salience rates candidate questions on depth and scientific utility. By asking the original authors to make this judgment, ratings reflect writers' intent, an aspect absent from most existing datasets. Figure usefulness and answerability help to determine if a question is a multimodal QUD: the figure has to be useful, and the figure with its caption must not already answer the question. 

Formally, a candidate question is multimodal QUD if: (1) the figure is useful, (2) the question is not answerable from figure and its caption (3) the answer generated from $(F, C_F, P_F)$ must be correct. 
As mentioned earlier, we keep the non-QUDs in the dataset (and marked), since they may be valuable as hard negative examples. 

\paragraph{Inter-annotator agreement.}
Since the authors themselves performed the annotation, we view their labels as inherently trustworthy. To further validate this point, for a subset of 40 questions, we recruited two co-authors from the same paper to independently annotate them (Table~\ref{tab:human-iaa}). The initial agreement values are moderate to high for most categories. We find that most disagreements largely reflect different rating strictness. After discussing the guidelines together, the annotators calibrated their thresholds.\footnote{For the ordinal dimensions we collapse labels to the operational categories: acceptable question grammar, salient, high answer quality, acceptable answer, and useful figure. Figure answerability is already binary.} After calibration, every dimension then reaches at least $0.90$. Our analyses and experiments use the calibrated labels.

\begin{table}[t]
\centering
\small
\setlength{\tabcolsep}{4pt}
\begin{tabular}{lrrrrr}
\toprule
& & \multicolumn{2}{c}{\textbf{Independent}} & \multicolumn{2}{c}{\textbf{Calibrated}} \\
\cmidrule(lr){3-4} \cmidrule(lr){5-6}
\textbf{Dimension} & \textbf{$N$} & \textbf{Agree.} & \textbf{$\kappa$}
                   & \textbf{Agree.} & \textbf{$\kappa$} \\
\midrule
Question grammar   & 40 & 100.0\% & 1.000 & 100.0\% & 1.000 \\
Salience           & 40 & 75.0\%  & 0.500 & 97.5\%  & 0.950 \\
Answer quality     & 38 & 86.8\%  & 0.736 & 94.9\%  & 0.898 \\
Answer correct     & 39 & 97.4\%  & 0.948 & 100.0\% & 1.000 \\
Figure usefulness  & 40 & 87.5\%  & 0.750 & 95.0\%  & 0.900 \\
Figure answerability & 40 & 67.5\% & 0.350 & 95.0\%  & 0.900 \\
\bottomrule
\end{tabular}
\caption{Inter-annotator agreement on 40 doubly annotated questions, before
(\textbf{Independent}) and after (\textbf{Calibrated}) one annotator aligned
their rating thresholds in a joint calibration pass. We report $\kappa$ from \citet{randolph2005free} due to skewed distributions.}
\label{tab:human-iaa}
\end{table}

\paragraph{Label distribution.}
Most dimensions are heavily skewed towards the positive category: the authors call $88\%$ of the figures useful, $90\%$ of the questions salient, and $84\%$ of the answers acceptable (Table~\ref{tab:annotation-dist}). Figure answerability is the exception and splits the data almost evenly, with $49\%$ of the questions answerable from the figure and caption alone. However, this category has the higher levels of disagreement with $\kappa=0.35$ (Table~\ref{tab:human-iaa}). 
Notably, we find that salience values tend to be higher for QUDs: 94\% of multimodal QUDs are salient, vs.\ 89\% of non-QUDs (counts in Table~\ref{tab:annotation-dist}).

\begin{figure}[t]
\centering
\includegraphics[width=0.9\linewidth]{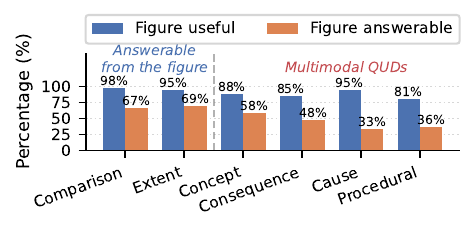}
\caption{Figure usefulness and figure answerability by question type. A large gap means the figure raises the question that cannot be answered by the figure itself. }
\label{fig:qud-clusters}
\end{figure}

\paragraph{Are certain question types more likely to be multimodal QUDs?}
\label{sec:annotation-analysis}
We analyze per-type rates of figure usefulness and figure answerability, using the expert annotations only.
Comparison and extent questions are more likely to be answered directly from the figure, meaning that a lower percentage of them are multimodal QUDs, in contrast to cause, consequence, procedural, and concept questions. 
The gap between figure usefulness and figure answerability (Figure~\ref{fig:qud-clusters}) is largest for causal questions, at 62 points. This shows that higher-level questions are more likely to be multimodal QUDs. 

\subsection{LLM-as-judge}\label{sec:llm-judge-validation}
We use GPT-5-mini as a zero-shot multimodal judge to score every question on the same seven dimensions as the experts, taking the figure together with the textual context and cited evidence used in annotation.
On the doubly-labeled questions, treating \texttt{yes} and \texttt{partial} as acceptable, the judge attains 89\% precision and $F_1 = 0.90$ on answer correctness.\footnote{The validation covers 778 matched pairs of expert and judge labels.} Agreement between LLM-as-judge and expert labels ranges from 64\% for salience to 93\% for question grammar (Table~\ref{tab:judge-iaa}).

\begin{table}[t!]
\centering
\small
\begin{tabular}{lrrr}
\toprule
\textbf{Dimension} & \textbf{Pairs} & \textbf{Agree.} & \textbf{$\kappa$} \\
\midrule
Question grammar & 786 & 93.1\% & 0.863 \\
Figure usefulness & 766 & 91.4\% & 0.828 \\
Answer correct & 778 & 81.9\% & 0.638 \\
Answer quality & 767 & 73.1\% & 0.463 \\
Figure answerability & 734 & 70.0\% & 0.401 \\
Salience & 779 & 63.8\% & 0.276 \\
Figure type & 720 & 63.5\% & 0.544 \\
\bottomrule
\end{tabular}
\caption{Human--judge agreement per dimension, with $\kappa$ from \citet{randolph2005free}.}
\label{tab:judge-iaa}
\end{table}

\section{Inquisitive Question Generation}
\label{sec:method}

This section presents experiments on multimodal question generation given only the figure $F$ and its prior context $C_F$. 
Formally, we produce questions $Q_F$:
\[Q_F \sim \pi_{LLM}(\cdot|F, C_F)\]
Unlike the \mqud{} construction (\S\ref{sec:generation}), this framework does not directly generate QUDs since the answer paragraphs $P_F$ are not provided and an answer is not guaranteed to be in the paper. Instead, this formulation explores the expectation space of $C_F$ and generates $Q_F$'s that satisfy other properties of a QUD, such as inquisitiveness and salience. 

We believe separating the QUD generation into these two steps is valuable for broader applications of this work: a model capable of generating inquisitive questions whose answers are not in the paper is useful to propose unexplored hypotheses and facilitate scientific discovery. In Section~\ref{sec:experiments}, we show that \mqud{} provides valuable supervision signals for such a vision. Future work on multimodal QUD parsing can also use this model with rejection sampling using a question answering module.

\subsection{Supervised fine-tuning}
\label{sec:sft}

We fine-tune Qwen3.5-9B with LoRA on filtered and augmented question generation data. Of the 592 human-annotated questions with acceptable answers (Table~\ref{tab:annotation-dist}), 468 also have useful figures and form the unaugmented SFT seed set. We augment the seeds with rephrased variants (Appendix~\ref{app:prompt-rephrase}) and hold out 51 items for validation. Only expert-annotated questions are used for SFT. We use GPT-5-mini to verify that augmenting the questions preserves the original answer. Training details are in Appendix~\ref{app:training}.

The Qwen generation model receives the trigger context, $C_F$, (title, abstract, figure, and caption) and predicts $Q_F$. In this setting, the model generates \emph{potential questions} since it does not have access to the paper body. 
We use a 51-item held-out diagnostic set for fine-grained loss analyses, and separately evaluate generalization on 9 papers never seen during training (Appendix~\ref{app:additional-tables}).

\section{Experiments}
\label{sec:experiments}
We evaluate whether training on \mqud{} increases the rate of salient questions that target the discourse role a figure plays in the paper.
Specifically, we test three hypotheses:
\begin{enumerate}[nosep, leftmargin=*]
    \item[\textbf{H1}] Content-specific visual grounding: the SFT model should be sensitive to which figure is presented, not just whether any figure is present. We test this by swapping figures within the same paper (\S\ref{sec:swap}).
    \item[\textbf{H2}] Figure dependency: After training on \mqud{}, the model should rely more on the figure when generating questions. We measure this with visual information gain (\S\ref{sec:grounding-results}).
    \item[\textbf{H3}] Question quality: The discourse framework present in \mqud{} should cause models to generate more salient questions after SFT. We assess this through quality evaluation of generated questions and provide a comparison to closed-sourced models (\S\ref{sec:llm-judge}).
\end{enumerate}

\begin{figure}[t]
\centering
\includegraphics[width=0.98\linewidth]{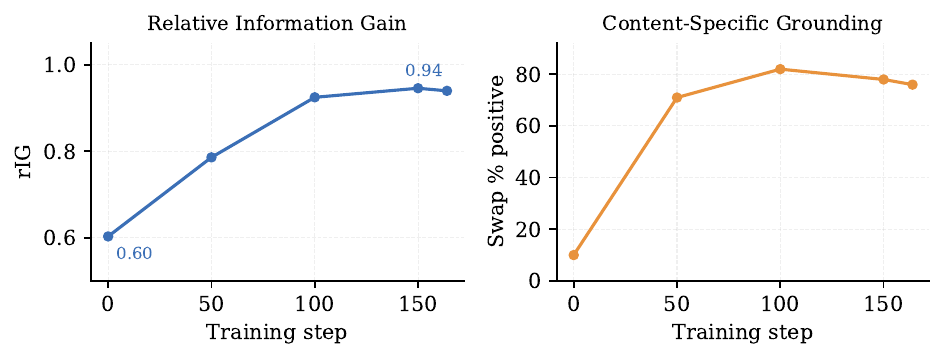}
\caption{rIG and content-specific grounding over training steps for our VLM trained on \mqud{}.} %
\label{fig:training-curves}
\end{figure}

\begin{table}[t]
\centering
\small
\setlength{\tabcolsep}{2pt}
\renewcommand{\arraystretch}{1.12}
\begin{tabular}{@{}lccc@{}}
\toprule
\textbf{Model} & \textbf{rIG} $\uparrow$ & \textbf{Swap} & \textbf{Swap \% pos} $\uparrow$ \\
\midrule
Base (zero-shot) &
\begin{tabular}[c]{@{}c@{}}0.60\\[-1pt]{\scriptsize[0.49, 0.73]}\end{tabular} &
\begin{tabular}[c]{@{}c@{}}$-$0.378\\[-1pt]{\scriptsize[$-$0.49, $-$0.27]}\end{tabular} &
11\% \\
\addlinespace[2pt]
+ SFT &
\begin{tabular}[c]{@{}c@{}}\textbf{0.94}\\[-1pt]{\scriptsize[0.69, 1.21]}\end{tabular} &
\begin{tabular}[c]{@{}c@{}}\textbf{$+$0.153}\\[-1pt]{\scriptsize[$+$0.04, $+$0.26]}\end{tabular} &
\textbf{71\%} \\
\bottomrule
\end{tabular}
\caption{Visual grounding diagnostics. rIG uses 51 diagnostic items; strict within-paper swap uses 38 same-paper figure swaps. Brackets: bootstrap 95\% CIs.}
\label{tab:grounding-results}
\end{table}

\subsection{Figure swap (H1)}\label{sec:swap}

\paragraph{Metric.}

This metric asks whether the model depends on \emph{this} figure rather than on any figure. Let $Q_F$ be the question written for a figure $F$, and let $F'$ denote a different figure from the same paper. $\mathcal{L}(Q_F \mid C_F, F)$ is the mean per-token negative log-likelihood the model assigns to $Q_F$ given the trigger context $C_F$ and the figure; $\mathcal{L}(Q_F \mid C_F)$ is the loss with no figure. The \textbf{swap gap} replaces $F$ with $F'$:
\begin{equation}
\Delta^{\text{swap}}_F = \mathcal{L}(Q_F \mid C_F, F') - \mathcal{L}(Q_F \mid C_F)
\label{eq:swap}
\end{equation}
A positive $\Delta^{\text{swap}}_F$ means a mismatched figure hurts more than no figure at all, which is what content-specific grounding predicts.

\begin{figure}[t]
\centering
\includegraphics[width=\linewidth]{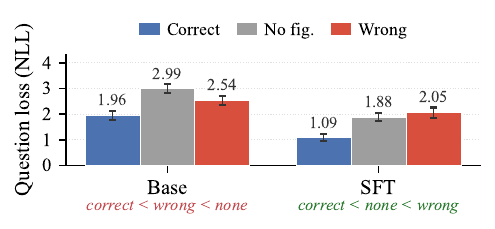}
\caption{Figure swap on the diagnostic items whose paper has a second figure. Lower loss is better. Base: the wrong figure still helps (correct $<$ wrong $<$ none). SFT: the wrong figure now hurts (correct $<$ none $<$ wrong), indicating content-specific grounding. Error bars: bootstrap 95\% CIs. }
\label{fig:swap-training}
\end{figure}

\paragraph{Results.} We compute the swap gap
on the 38 of 51 diagnostic items whose paper has a second figure; the remaining 13 items come from papers with a single qualifying figure.
Before SFT, even an incorrect figure $F'$ lowers question's NLL relative to no figure ($\Delta^{\text{swap}}_F < 0$), indicating a generic bias toward visual input. After SFT, a wrong figure \emph{increases} NLL beyond the text-only baseline ($\Delta^{\text{swap}}_F > 0$), indicating reliance on specific figure content (Table~\ref{tab:grounding-results}, Figure~\ref{fig:swap-training}). Negative swap values after SFT are concentrated on comparison and procedural question types, where figures from the same paper share a similar visual structure. 

To test generalization, we also train a model that never sees the held-out papers (\S\ref{sec:sft}) and evaluate on them. The swap-positive rate reaches 74\% indicating that grounding transfers to unseen papers. Appendix~\ref{app:frozen-scorer} repeats the swap with a frozen scorer and adds a retrieval check.

\subsection{Visual information gain (H2)}
\label{sec:grounding-results}

\paragraph{Metric.}
We measure figure dependency by asking how much harder the reference question is to predict when the figure is removed. We evaluate under two conditions: \textbf{multimodal} (mm), where the model receives $C_F$ and $F$, and \textbf{text-only} (to), where $F$ is withheld. $\mathcal{L}_{\text{mm}}$ and $\mathcal{L}_{\text{to}}$ denote the loss for the multimodal and text-only cases, respectively. We define \textbf{visual information gain} as
\[
\Delta\mathcal{L}_F = \mathcal{L}_{\text{to}}(Q_F) - \mathcal{L}_{\text{mm}}(Q_F),
\]
and report a \textbf{relative information gain} (rIG):
\begin{equation}
  \text{rIG}_F = \frac{\mathcal{L}_{\text{to}}(Q_F) - \mathcal{L}_{\text{mm}}(Q_F)}{\mathcal{L}_{\text{mm}}(Q_F)}
  \label{eq:rig}
\end{equation}
so that complex questions do not automatically appear more figure-dependent (see Appendix~\ref{app:calibration}). The unnormalized loss difference is a model-based estimate of the conditional information contributed by $F$ when averaged over reference QUDs. Because finite-model calibration also affects this difference, we report the normalized rIG.

\paragraph{Results.}
After training on multimodal QUDs, the model relies more on the figure when generating questions. rIG increases from 0.60 to 0.94 ($p < 0.0001$; Table~\ref{tab:grounding-results}); per-type values are reported descriptively in Appendix~\ref{app:info-gain-detail}. A text-only SFT ablation achieves a comparable swap rate but much lower rIG, confirming that the two diagnostics capture distinct model properties: swap can reflect discourse-structural patterns, while rIG measures sensitivity to the provided visual input (Table~\ref{tab:ablations}).

\subsection{Question quality (H3)}
\label{sec:llm-judge}
\label{sec:main-results}

\paragraph{Metric.} We evaluate question quality using
the LLM judge presented in \S\ref{sec:llm-judge-validation} to score each question on \textbf{Salience}, \textbf{Figure usefulness}, and \textbf{Figure answerability}.

\paragraph{Results.}
On the held-out papers, the judge scores the reference questions, i.e., the filtered candidates in \mqud{} (\S\ref{sec:generation}), and each model's generations (Table~\ref{tab:quality}). Before SFT, over half of the generated questions are answerable from the figure and caption alone, indicating that the model reads the figure rather than questioning it. After SFT, questions go beyond the figure nearly as often as the references and score higher on salience, second only to GPT-4o. Half of the reference questions meet both figure conditions; SFT comes closest at 40\% (examples in \S\ref{app:qualitative-grounding}).

\begin{table}[t]
\centering
\small
\setlength{\tabcolsep}{3pt}
\begin{tabular}{@{}lccc@{}}
\toprule
\textbf{Source} & \textbf{Salient} & \textbf{Fig.\ useful} & \textbf{Not ans.\ by fig.} \\
\midrule
Base (zero-shot) & 56.7 & 90.0 & 45.6 \\
\textbf{+ SFT}   & \textbf{62.2} & \textbf{84.4} & \textbf{53.3} \\
\midrule
GPT-4o           & 73.3 & 86.7 & 54.4 \\
GPT-5-mini       & 52.2 & 86.7 & 43.3 \\
\addlinespace[2pt]
Reference        & 51.1 & 84.4 & 57.8 \\
\bottomrule
\end{tabular}
\caption{Question quality on the held-out papers: share of questions the \S\ref{sec:llm-judge-validation} judge rates salient (yes or somewhat), figure-useful, and not answerable from the figure and caption alone.}
\label{tab:quality}
\end{table}

\subsection{Summary of findings}
\label{sec:summary_of_findings}
\textbf{All three hypotheses are supported.} For H1, the base model exhibits a general bias toward image content. A figure that $Q_F$ was not written for decreases the NLL compared to no figure. After training on \mqud{}, the swap-positive rate increases from $11\%$ to $71\%$, showing that the model is sensitive to figure-specific content. For H2, relative information gain rises from $0.60$ to $0.94$ after fine-tuning, indicating that the model relies more on the figure when generating questions. For H3, over half of the base model's questions are answerable from the figure and caption alone. Training on \mqud{} raises salience from $56.7\%$ to $62.2\%$ and the share not answerable from the figure alone from $45.6\%$ to $53.3\%$, close to the reference rate of $57.8\%$. Together these results show that \mqud{} is a valuable training resource for generating inquisitive questions that are salient to a paper's research goals and cannot be derived from simple chart retrieval.

\section{Conclusion}
\label{sec:discussion}

Text-only QUD models discourse as questions raised and resolved by sentences. Extending QUD to the multimodal setting allows figures to participate in this relation. Some figure-triggered questions are answered directly by the figure; others require the reader to connect a visual observation to a textual explanation. The six primary question types differ in the prevalence of these two item-level dependency patterns, but no type determines a question's dependency label.
This structure suggests a path toward generating deeper scientific questions. Rather than asking only what a figure shows, a system can ask grounded questions about why a pattern occurs or how it relates to the paper's argument.
We showed that \mqud{} provides valuable training data for this task, generating salient, grounded questions suitable for scientific research.

\section*{Limitations}
Our salience labels reflect author intent, not questions independently produced by unseen readers. The task requires annotators who authored the paper, and recruiting several such co-authors is rarely possible, so most judgments rest on a single expert. The dataset covers 56 papers in three domains and requires LaTeX source, so broader fields and non-LaTeX documents remain untested. 

Our design choices pose some limitations on question diversity, which we adopt for practical reasons: GPT-4o sees $P_F$ when proposing candidates; although the prompt separates trigger context from evidence, this does not causally establish that $P_F$ had no effect on question choice. Additionally, questions whose answers fall out of $P_F$ are not in the current dataset. 

Finally, for evaluation, rIG and figure swap diagnose model behavior; they do not by themselves validate question quality, and within-paper swaps may be less discriminative when figures are visually similar.

\nocite{dataset1712_08485v2,dataset1901_11429v2,dataset2005_04762v3,dataset2010_02494v1,dataset2108_07344v2,dataset2112_13864v2,dataset2209_06315v1,dataset2209_06687v3,dataset2210_01963v4,dataset2212_11301v2,dataset2301_08968v2,dataset2304_02662v1,dataset2305_13486v1,dataset2305_19358v3,dataset2306_00738v2,dataset2307_05435v4,dataset2307_14991v2,dataset2309_11415v1,dataset2310_00198v2,dataset2311_18129v1,dataset2312_03066v2,dataset2405_00955v1,dataset2405_05506v2,dataset2405_14619v3,dataset2406_16320v3,dataset2406_17947v2,dataset2410_07951v1,dataset2410_22590v2,dataset2411_19332v1,dataset2412_12072v2,dataset2501_07391v1,dataset2502_03201v1,dataset2502_12317v1,dataset2502_15969v4,dataset2502_19391v1,dataset2502_19395v1,dataset2504_09373v2,dataset2505_14963v1,dataset2505_17127v2,dataset2505_20063v2,dataset2506_03902v1,dataset2506_17476v1,dataset2507_03165v1,dataset2507_03239v1,dataset2507_13328v2,dataset2508_13650v2,dataset2509_21606v2,dataset2510_01322v1,dataset2510_24538v1,dataset2511_08598v1,dataset2602_05902v1,dataset2602_19217v1,dataset2602_19224v1}
\section*{Acknowledgments}
We are grateful for all researchers who participated in our data collection process.
This work was partially supported by the U.S.\ National Science Foundation (NSF) under Cooperative Agreement 2421782
and the Simons Foundation grant MPS-AI-00010515 awarded to the NSF-Simons AI Institute for Cosmic Origins,\footnote{\url{https://www.cosmicai.org/}} the NSF AI Institute for Foundations of Machine Learning (IFML), NSF grant 2019844, as well as NSF CAREER award IIS-2145479.

\bibliography{references,dataset_papers_canonical}

\clearpage
\appendix

\section{Calibration decomposition}
\label{app:calibration}

We provide a formal justification for why raw $\Delta\mathcal{L}$ can decrease after training even when figure dependency increases. Any cross-entropy loss can be decomposed as $\mathcal{L} = H + \mathrm{KL}$, where $H$ is the true entropy and $\mathrm{KL}$ is the gap between the model and the true distribution. Applying this to both conditions:
\begin{equation}
  \Delta\mathcal{L} \;=\; I(Q_F;\,F\mid C_F) \;+\; \bigl[\mathrm{KL}_{\text{to}} - \mathrm{KL}_{\text{mm}}\bigr]
\end{equation}
SFT minimizes cross-entropy on multimodal input, which can reduce $\mathrm{KL}_{\text{mm}}$ (the model's calibration gap with the figure) more than $\mathrm{KL}_{\text{to}}$ (calibration without the figure). This drives $\Delta\mathcal{L}$ \emph{up} even if the true figure dependency $I(Q_F;\,F\mid C_F)$ is unchanged. We therefore report rIG $= \Delta\mathcal{L}/\mathcal{L}_{\text{mm}}$, which normalizes by question difficulty so that harder questions do not dominate the metric. rIG captures the relative contribution of the figure more faithfully than raw $\Delta\mathcal{L}$.

\section{Additional grounding checks}
\label{app:frozen-scorer}
\label{app:additional-tables}
The figure swap of \S\ref{sec:swap} scores the model with its own loss; here we rerun it with a separate frozen model as the scorer.

\begin{figure}[h]
\centering
\includegraphics[width=\linewidth]{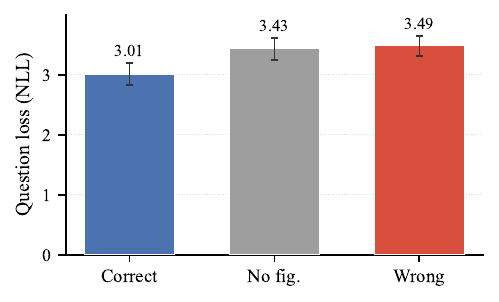}
\caption{Question NLL for the reference questions on the held-out papers under three figure conditions, scored by the frozen Qwen2.5-VL-32B.}
\label{fig:swap-disjoint}
\end{figure}

\paragraph{Loss diagnostics.}
The frozen scorer, Qwen2.5-VL-32B-Instruct, reproduces the pattern of \S\ref{sec:swap} on the held-out papers: the correct figure scores best, and the wrong figure hurts most (content gap $\mathcal{L}_{\text{swap}} - \mathcal{L}_{\text{mm}} = +0.47$; Figure~\ref{fig:swap-disjoint}).\footnote{Content gap $+0.471$ $[0.371, 0.581]$, bootstrap 95\% CI.}

\paragraph{Retrieval check.}
We also test retrieval: a frozen SigLIP2 encoder takes each question as the query and ranks all 243 corpus figures. Reference questions retrieve their own figure far above chance (41\% vs.\ 0.4\%), and generated questions do so even more often, indicating that the questions describe their specific figure rather than figures in general.

\begin{table*}[t]
\centering
\small
\begin{tabular}{lccc}
\toprule
\textbf{Model} & \textbf{rIG} & \textbf{Swap} & \textbf{Swap \% pos} \\
\midrule
Qwen3-VL-8B Base (zero-shot) & --- & $-$0.702 & 8\% \\
Qwen3.5-9B + SFT (text-only trained) & 0.27 & $+$0.129 & 74\% \\
Qwen3.5-9B + SFT (shuffled images) & --- & $+$0.024 & 63\% \\
Qwen3-VL-8B-Instruct (few-shot) & 1.15 & --- & 24\% \\
\bottomrule
\end{tabular}
\caption{Ablation results. rIG is computed on the diagnostic set; strict within-paper swap metrics use the 38 items with another available figure from the same paper. Text-only SFT reaches a comparable swap rate to image-conditioned SFT (74\% vs.\ 71\%) but much lower rIG (0.27 vs.\ 0.94), indicating that high swap alone can be achieved from textual discourse patterns without visual grounding. Shuffled-image SFT shows reduced swap (63\%), confirming correct image-text pairing matters. Qwen3-VL base confirms the finding generalizes across architectures.}
\label{tab:ablations}
\end{table*}

\section{Per-type visual information gain}
\label{app:info-gain-detail}

We report per-type rIG for the generation model in Table~\ref{tab:info-gain-app}. rIG increases after SFT for comparison and extent, and cause shows the largest increase; concept is the only type without an increase.

\begin{table}[h]
\centering
\small
\begin{tabular}{lrcc}
\toprule
\textbf{Question type} & \textbf{$N$} & \textbf{Base rIG} & \textbf{SFT rIG} \\
\midrule
Extent & 10 & 0.56 & \textbf{0.74} \\
Comparison & 10 & 0.43 & \textbf{0.77} \\
Cause & 13 & 0.78 & \textbf{1.54} \\
Consequence & 8 & 0.66 & \textbf{1.01} \\
Procedural & 5 & 0.34 & \textbf{0.57} \\
Concept & 5 & 0.58 & 0.55 \\
\midrule
Overall & 51 & 0.60 & \textbf{0.94} \\
\bottomrule
\end{tabular}
\caption{Per-type rIG on the diagnostic set.}
\label{tab:info-gain-app}
\end{table}

\section{Dataset comparison}
\label{app:dataset-comparison}

\begin{figure}[h]
\centering
\includegraphics[width=0.85\linewidth]{figures/chartqa_example.png}\\[2pt]
{\small \textbf{ChartQA}: \emph{``What is the difference in value between Lamb and Corn?''}}\\[8pt]
\includegraphics[width=0.5\linewidth]{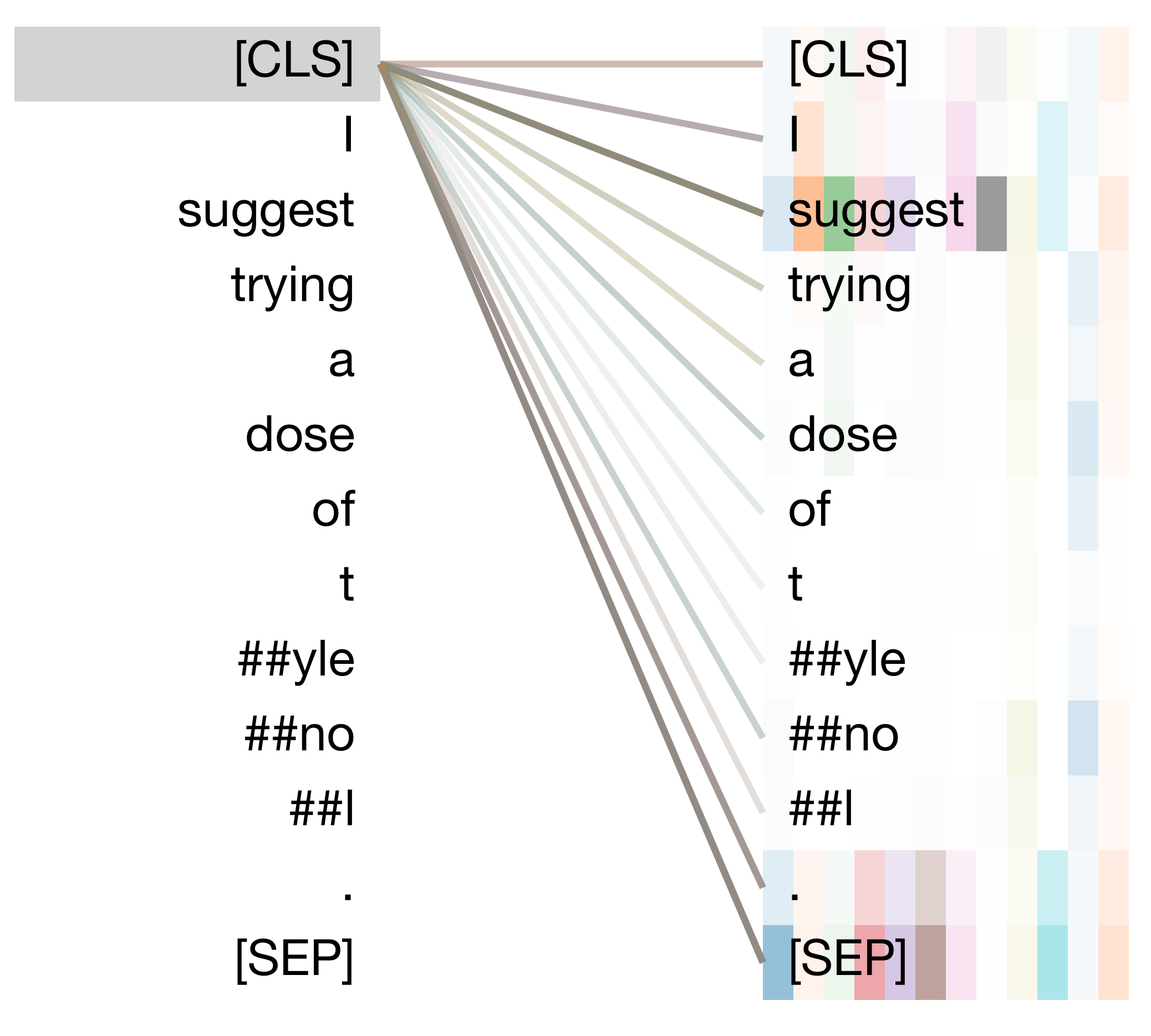}\\[2pt]
{\small \textbf{\mqud{}}: \emph{``Why does the model assign higher attention weights to the word `suggest' in this sentence?''}}
\caption{Representative examples from ChartQA and \mqud{}. ChartQA asks for a numeric difference readable from the chart. \mqud{} asks why a pattern exists, requiring integration of the attention visualization with the paper's text.}
\label{fig:dataset-examples}
\end{figure}

Figure~\ref{fig:dataset-examples} shows representative examples from ChartQA and \mqud{}: the ChartQA question extracts a value from the chart, while the \mqud{} question asks for the cause of the observed pattern.

\section{Dataset examples}
\label{app:examples}

We present six examples from \mqud{}, one per primary question type. Each example includes its stored answer and a short excerpt of the source evidence. All are rated figure-useful.

\begin{prompt}[title={Example 1: Comparison question (answered by the figure)}]
\textbf{Paper:} \emph{Decomposing Generalization: Models of Generic, Habitual, and Episodic Statements}\\[4pt]
\begin{center}
\includegraphics[width=0.45\linewidth]{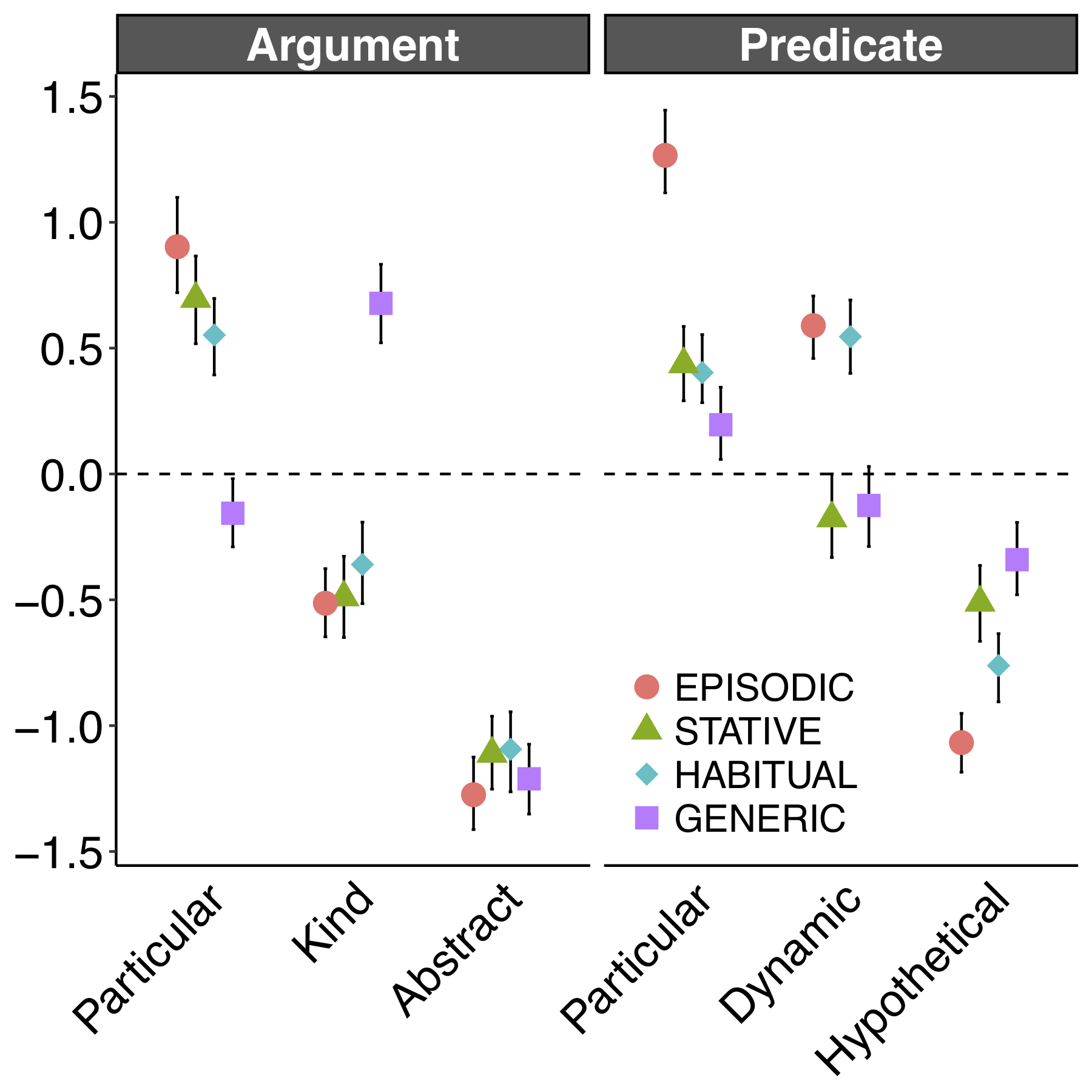}
\end{center}
\textbf{Question:} \emph{How do the tendencies of arguments to be kind-referring or particular-referring compare in the figure?}\\
\textbf{Answer:} Arguments have a slight tendency to refer to either a kind or a particular ($\rho{=}-0.33$). `Place' falls in the lower right quadrant and `transportation' in the upper left.\\[4pt]
\textbf{Source evidence:} \emph{Arguments show a slight tendency ($\rho{=}-0.33$) to refer to either a kind or a particular; the text identifies `place' and `transportation' as examples in opposite quadrants.}\\[4pt]
\textbf{Annotation:} figure-useful=useful, answered-by-figure=\colorbox{green!20}{yes}, salience=salient, answer-correct=acceptable, answer-quality=high, figure-type=data, q-grammar=acceptable\\
\end{prompt}

\begin{prompt}[title={Example 2: Cause QUD (multimodal QUD)}]
\textbf{Paper:} \emph{Help! Need Advice on Identifying Advice}\\[4pt]
\begin{center}
\includegraphics[width=0.3\linewidth]{figures/example_attention.png}
\end{center}
\textbf{Question:} \emph{Why does the model assign higher attention weights to the word `suggest' in this sentence?}\\
\textbf{Answer:} The model assigns higher attention to `suggest' because it is a key lexical cue indicative of advice. The visualization confirms that the model identifies advice-related words even when the sentence was incorrectly annotated.\\[4pt]
\textbf{Source evidence:} \emph{The paper states that the model attends to `suggest,' correctly predicts advice despite a noisy label, and latches onto lexical cues indicative of advice.}\\[4pt]
\textbf{Annotation:} figure-useful=useful, answered-by-figure=\colorbox{red!20}{no}, salience=salient, answer-correct=acceptable, answer-quality=high, figure-type=result, q-grammar=acceptable\\
\end{prompt}

\begin{prompt}[title={Example 3: Extent question (answered by the figure)}]
\textbf{Paper:} \emph{Stable Anisotropic Regularization}\\[4pt]
\begin{center}
\includegraphics[width=0.55\linewidth]{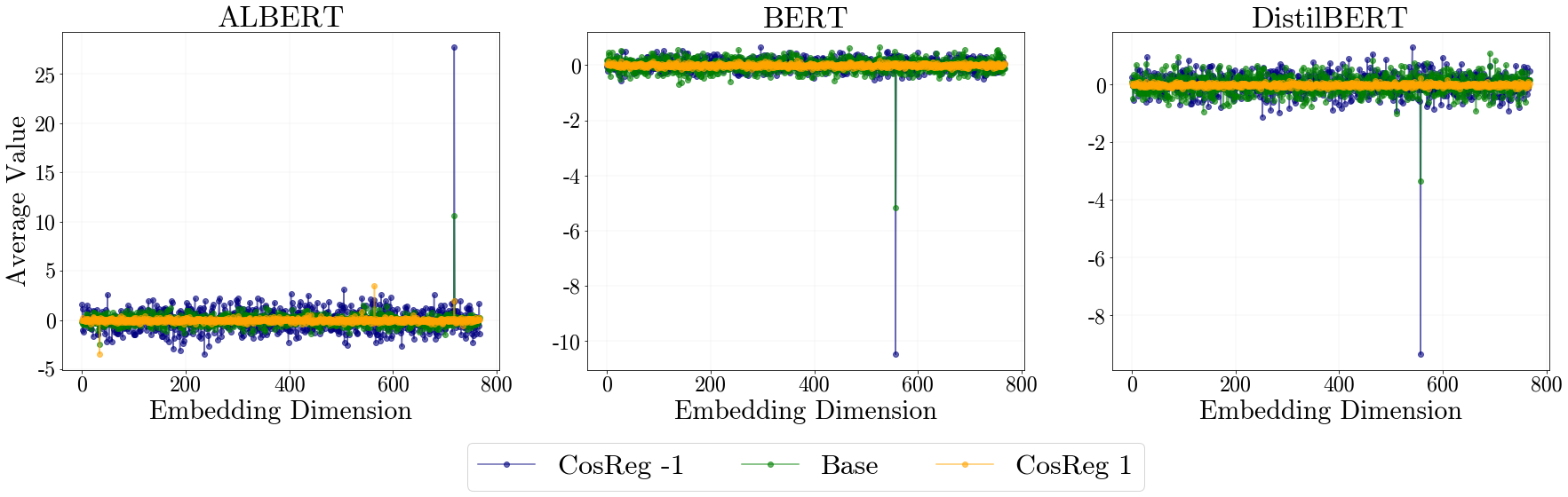}
\end{center}
\textbf{Question:} \emph{To what extent does the application of CosReg -1 affect the mean values in the embedding dimensions for DistilBERT?}\\
\textbf{Answer:} The CosReg setting of $-1$ increases the representation mean, with a particularly large change in one dimension, but does not improve isotropy.\\[4pt]
\textbf{Source evidence:} \emph{The paper reports that CosReg alters the mean of model activations, particularly in one dimension, while leaving representation isotropy unimproved.}\\[4pt]
\textbf{Annotation:} figure-useful=useful, answered-by-figure=\colorbox{green!20}{yes}, salience=salient, answer-correct=acceptable, answer-quality=high, figure-type=result, q-grammar=acceptable\\
\end{prompt}

\begin{prompt}[title={Example 4: Consequence QUD (multimodal QUD)}]
\textbf{Paper:} \emph{How People Talk about Each Other: Modeling Generalized Intergroup Bias and Emotion}\\[4pt]
\begin{center}
\includegraphics[width=0.55\linewidth]{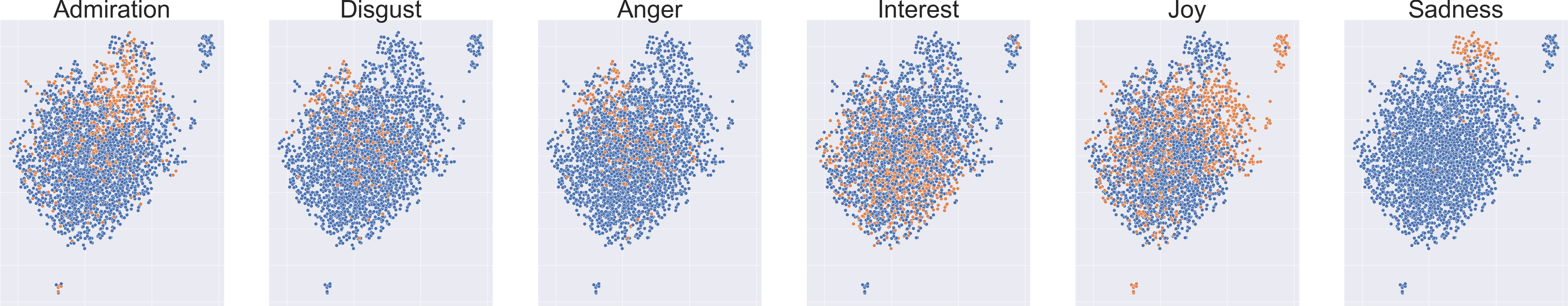}
\end{center}
\textbf{Question:} \emph{What are the implications of the overlap in tweet representations for modeling interpersonal group relationships?}\\
\textbf{Answer:} Interpersonal emotions do not form fully separated regions, although some opposite emotions are moderately separable. The paper interprets this as evidence that emotions encode some topic- or domain-level properties of tweets.\\[4pt]
\textbf{Source evidence:} \emph{The paper notes substantial overlap across emotion labels, moderate separation for pairs such as admiration/disgust and joy/sadness, and concludes that emotions define some topic- or domain-level properties.}\\[4pt]
\textbf{Annotation:} figure-useful=useful, answered-by-figure=\colorbox{red!20}{no}, salience=salient, answer-correct=acceptable, answer-quality=high, figure-type=data, q-grammar=acceptable\\
\end{prompt}

\begin{prompt}[title={Example 5: Procedural QUD (multimodal QUD)}]
\textbf{Paper:} \emph{Multilingual Code Co-Evolution Using Large Language Models}\\[4pt]
\begin{center}
\includegraphics[width=0.55\linewidth]{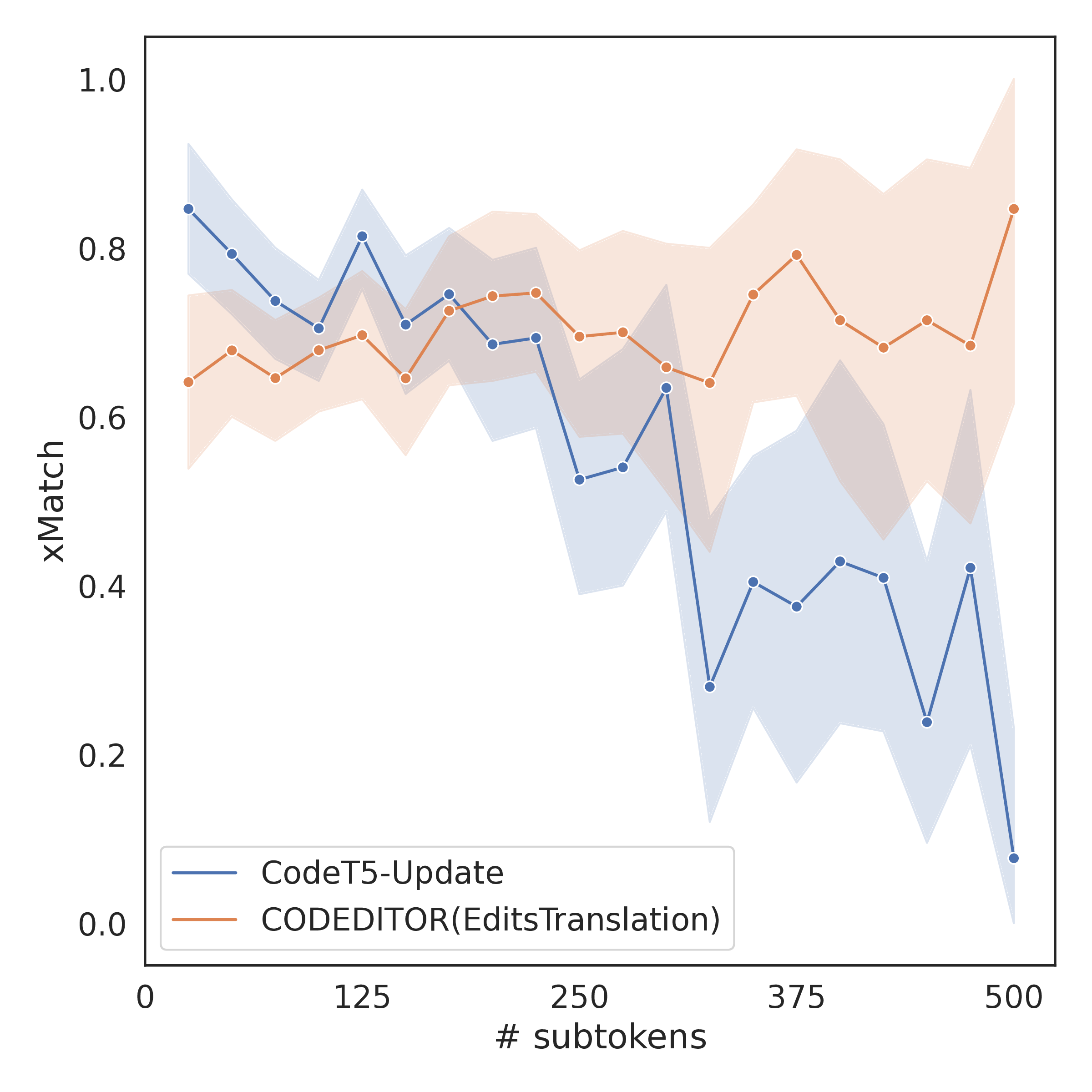}
\end{center}
\textbf{Question:} \emph{By what mechanism does CODEDITOR maintain stable performance across varying numbers of subtokens?}\\
\textbf{Answer:} CODEDITOR maintains stable performance by modeling code changes as edit sequences, which allows it to focus on transforming specific edits rather than generating entire code segments. This approach is particularly effective for handling longer input sequences.\\[4pt]
\textbf{Source evidence:} \emph{The paper attributes stable performance on longer inputs to transforming edits instead of generating the entire new code, as CodeT5-Update does.}\\[4pt]
\textbf{Annotation:} figure-useful=useful, answered-by-figure=\colorbox{red!20}{no}, salience=salient, answer-correct=acceptable, answer-quality=high, figure-type=result, q-grammar=acceptable\\
\end{prompt}

\begin{prompt}[title={Example 6: Concept QUD (multimodal QUD)}]
\textbf{Paper:} \emph{exLong: Generating Exceptional Behavior Tests with Large Language Models}\\[4pt]
\begin{center}
\includegraphics[width=0.55\linewidth]{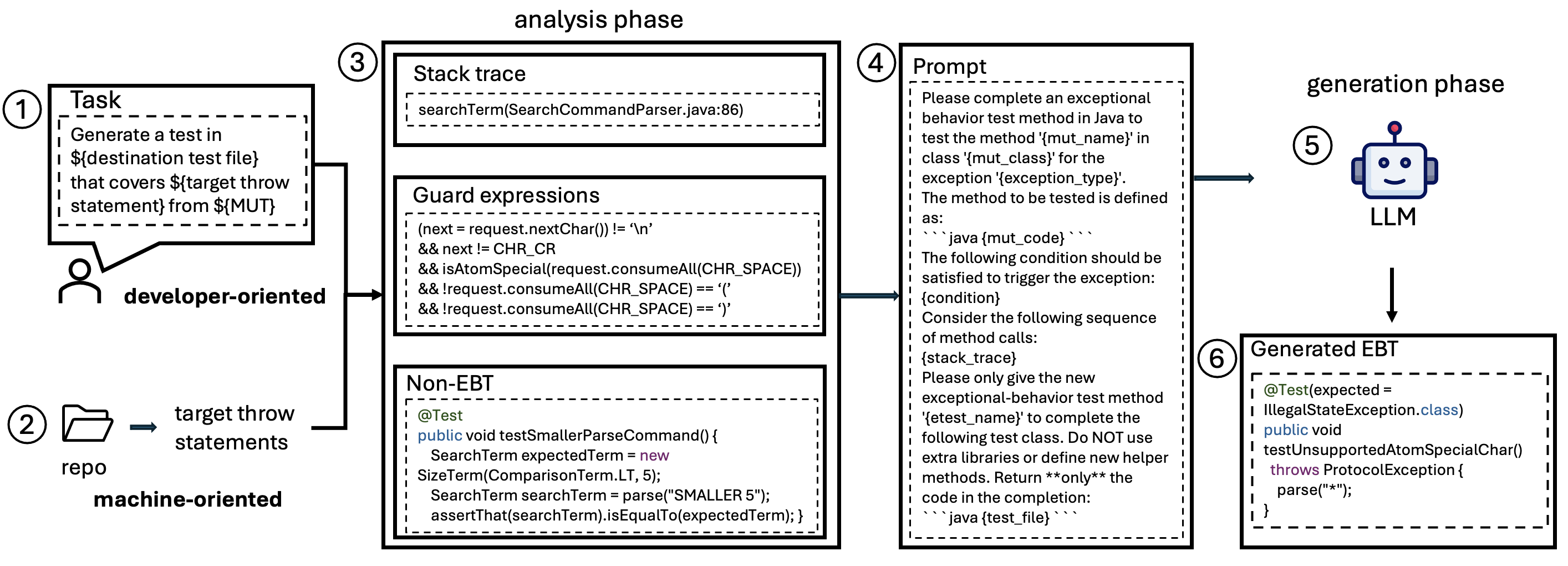}
\end{center}
\textbf{Question:} \emph{What is the significance of the generated EBT shown in box 6 compared to the non-EBT in box 3?}\\
\textbf{Answer:} Box 3 supplies analysis context, including a guard expression and an existing non-EBT, while box 6 is the LLM-generated test that targets the requested exception. The contrast illustrates how exLong turns program-analysis context into an exceptional-behavior test.\\[4pt]
\textbf{Source evidence:} \emph{The paper defines guard expressions as constraints required to follow a trace and states that the model receives this collected context to produce an exceptional-behavior test.}\\[4pt]
\textbf{Annotation:} figure-useful=useful, answered-by-figure=\colorbox{red!20}{no}, salience=salient, answer-correct=acceptable, answer-quality=high, figure-type=method, q-grammar=acceptable\\
\end{prompt}

\section{Extended dataset analysis}
\label{app:dataset-analysis}

The analyses below use the full annotation set, including the LLM-judge labels, before any training-data filtering. For training, we further filter to 468 questions where annotators rated both the answer as correct and the figure as useful, then augment with rephrased variants to obtain 1,308 training examples (\S\ref{sec:sft}).

\begin{figure}[h]
\centering
\includegraphics[width=\linewidth]{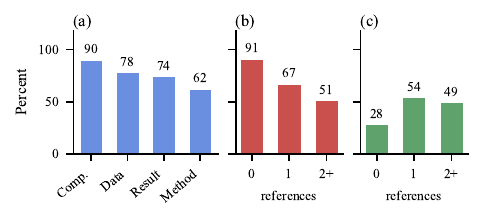}
\caption{Dataset annotation properties. (a) Figure type vs.\ usefulness. (b) Reference frequency vs.\ usefulness. (c) Reference frequency vs.\ answer quality.}
\label{fig:refcount}
\end{figure}

We analyze how figure properties relate to annotation dimensions (Figure~\ref{fig:refcount}).

\textbf{Figure type shapes usefulness.} Comparison figures are most visually useful at 90\%, while method figures are least useful at 62\% (panel a). We interpret this as reflecting inherent function: comparison figures exist to enable visual comparison, whereas method figures often illustrate what the text already describes.

\textbf{More-referenced figures are less visually useful.} We measure the number of distinct paper sections that cite each figure. Figures referenced in more sections are rated \emph{less} useful ($\rho = -0.24$, $p < 10^{-5}$; panel b). When the text discusses a figure extensively, the figure becomes supplementary; when the text says little, the figure carries the information.

\textbf{More-referenced figures yield better answers.} Despite being less useful, questions about frequently-referenced figures receive higher answer quality ratings ($\rho = 0.16$, $p = 0.005$; panel c). We attribute this to richer textual context for answer generation.

\section{Annotation details}
\label{app:annotation}

\begin{table*}[t]
\centering
\small
\begin{tabular}{llp{9cm}}
\toprule
\textbf{Stage} & \textbf{Actor} & \textbf{Role} \\
\midrule
Expert annotation & paper authors & label seven dimensions; may rewrite questions \\
Section selection & GPT-4o-mini & picks the results and analysis sections whose figures we use \\
Candidate generation & GPT-4o & proposes candidate question--answer pairs \\
Label expansion \& augmentation & GPT-5-mini & labels unannotated questions; checks rephrased variants \\
Fine-tuned generator & Qwen3.5-9B & LoRA-trained question generation model \\
External scorer & Qwen2.5-VL-32B & frozen; scores the grounding diagnostics \\
\bottomrule
\end{tabular}
\caption{Model and annotation roles across the pipeline.}
\label{tab:pipeline-roles}
\end{table*}

\paragraph{Human--judge reliability.}
We match each available expert annotation to the GPT-5-mini rating of the same question after applying adjudicated corrections. Repeated expert annotations remain separate pairs, so the 778 answer-correct pairs correspond to 694 unique questions. Table~\ref{tab:judge-iaa} reports agreement and $\kappa$ per dimension. We select SFT originals from expert labels; the judge is used to label the non-expert subset and to verify rephrased variants.

\section{Blind A/B check of the judge}
\label{app:blind-ab}

\begin{figure}[h]
\centering
\includegraphics[width=0.9\linewidth]{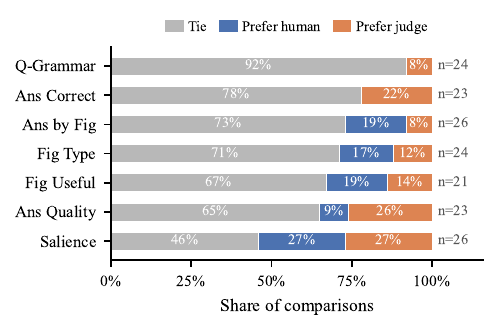}
\caption{Blind A/B sanity check of the LLM judge. One expert compared anonymized human and LLM judge ratings. Ties dominate all dimensions (46--92\%), with no significant preference for either source.}
\label{fig:blind-eval-main}
\end{figure}

One expert compared the human and the judge rating of the same question, with the source of each hidden, and selected the preferred rating or a tie on each of the seven dimensions. Ties dominate every dimension (46--92\%; Figure~\ref{fig:blind-eval-main}), indicating that the judge rates questions much as the experts do. This is a single-rater check, not an equivalence test; the per-dimension agreement in Appendix~\ref{app:annotation} remains the primary validation.

\section{Prompt templates}
\label{app:prompts}

\subsection{QUD generation prompt}
\label{app:prompt-generation}

The following prompt generates multimodal QUDs for a given figure, conditioned on the paper abstract, figure image, caption, and anchor paragraphs. It instructs GPT-4o to produce questions distributed across the six QUD types (\S\ref{sec:qud-types}).

\begin{prompt}[title={Prompt 1: Multimodal QUD Generation}]
Generate natural, research-oriented questions that arise from viewing this figure and the paper abstract, where answers are found in the paper text.\\

\textbf{Scenario:} An inquisitive researcher who read the abstract of this paper views this figure and its caption. What would they naturally wonder about what is shown in the figure?\\

\textbf{Critical requirements:}
\begin{enumerate}[nosep, leftmargin=*]
  \item \textbf{Questions arise from viewing the figure:} Reference what is \emph{visible} (trends, differences, patterns, values, annotations). Express genuine curiosity (why, how, what). Cannot be answered from the caption alone.
  \item \textbf{Answers from paper text:} 2--4 substantive sentences providing interpretation, cause, or context. Should provide insight, not just restate the question.
  \item \textbf{Natural research language:} Write from a researcher's perspective. Match the paper's sophistication level. Vary question structures naturally.
  \item \textbf{Specific to this figure:} Reference concrete visual elements (lines, bars, regions, numerical values). Use terms from the provided paragraphs, not generic placeholders.
\end{enumerate}
~\\
\textbf{Question type examples} (good \checkmark{} and bad \texttimes{}):
\begin{itemize}[nosep, leftmargin=*]
  \item \textbf{Cause:} \checkmark{} ``Why does the accuracy drop sharply after 100 tokens in the left panel?''
  \item \textbf{Comparison:} \checkmark{} ``How does the baseline's behavior differ from the proposed method in the high-resource setting?''
  \item \textbf{Extent:} \checkmark{} ``To what extent does the model's perplexity improve with additional training data?''
  \item \textbf{Consequence:} \checkmark{} ``What happens to the loss curve when the learning rate exceeds 0.01?''
  \item \textbf{Procedural:} \checkmark{} ``How does the model achieve the performance plateau visible around epoch 50?''
  \item \textbf{Concept:} \checkmark{} ``What does the shaded region in the time series represent in terms of model uncertainty?''
  \item[] \textbf{Generic failures for every type:} \texttimes{} ``What does the figure show?'' (too generic) \quad \texttimes{} ``What is the value of the blue line at x=5?'' (trivial)
\end{itemize}
~\\
\textbf{Context:} Paper: \texttt{\{paper\_name\}}. Abstract: \texttt{\{abstract\}}. Figure \texttt{\{figure\_number\}}, caption: \texttt{\{caption\}}, referenced \texttt{\{reference\_count\}} times. Other figures in paper: \texttt{\{other\_figures\}}. Text discussing this figure: \texttt{\{paragraphs\}}.\\

\textbf{Output:} JSON array with fields: \texttt{question}, \texttt{answer}, \texttt{answer\_source}, \texttt{question\_type} $\in$ \{cause, comparison, extent, consequence, procedural, concept\}, \texttt{difficulty} $\in$ \{medium, hard\}.
\end{prompt}

\subsection{Rephrase augmentation prompt}
\label{app:prompt-rephrase}

To diversify training data (\S\ref{sec:sft}), we generate structurally distinct rephrasings of each QUD--answer pair.

\begin{prompt}[title={Prompt 2: Rephrase Augmentation}]
Your goal is to produce DIVERSE rephrases that a real researcher might ask --- not just synonym swaps.

Given a question--answer pair about a scientific paper figure, generate \texttt{\{n\_variants\}} rephrased variants. Each variant MUST be structurally different from the others.\\

\textbf{Diversity requirements (critical):}
\begin{itemize}[nosep, leftmargin=*]
  \item \textbf{Variant 1:} Restructure the question substantially (e.g., change from direct to embedded question, split into sub-questions, or approach from a different angle while asking about the same thing). Answer should be CONCISE (1--2 sentences, $\sim$20--30 words).
  \item \textbf{Variant 2:} Use a notably different framing or perspective (e.g., if original asks ``why does X increase'', rephrase as ``what explains the upward trend in X''). Answer should be DETAILED (4--5 sentences, $\sim$60--80 words) with context and implications.
  \item Do NOT just swap synonyms. ``Why does X'' $\to$ ``What causes X'' is NOT enough variation.
  \item The two variants MUST have noticeably different answer lengths and structures.
\end{itemize}
~\\
\textbf{Preservation:} Same factual content, same figure references, same question type intent (\texttt{\{question\_type\}}).\\

Figure caption: \texttt{\{caption\}}. Source excerpt: \texttt{\{source\_excerpt\}}.\\
Original question: \texttt{\{question\}}. Original answer: \texttt{\{answer\}}.\\

\textbf{Output:} JSON array of objects with \texttt{question} and \texttt{answer} fields.
\end{prompt}

\subsection{Answer grounding check}
\label{app:prompt-grounding}

We use GPT-5-mini to verify that the answer still comes from the source text after rephrasing.

\begin{prompt}[title={Prompt 3: Answer Grounding Check}]
Check whether an answer to a scientific question is grounded in the provided text.\\

\textbf{Criteria:}
\begin{enumerate}[nosep, leftmargin=*]
  \item Every claim in the answer must be traceable to the caption and/or source text.
  \item The answer must be concrete and informative --- non-answers like ``can be identified by looking at the figure'' are \textbf{not} grounded.
\end{enumerate}
~\\
Figure caption: \texttt{\{caption\}}. Source text: \texttt{\{source\_text\}}.\\
Question: \texttt{\{question\}}. Answer: \texttt{\{answer\}}.\\

\textbf{Output:} JSON with \texttt{grounded} (boolean) and \texttt{reason} (brief explanation).
\end{prompt}

\subsection{Zero-shot LLM judge}
\label{app:prompt-judge}

We evaluate all 1,250 questions using a zero-shot LLM judge (GPT-5-mini) that scores on the same seven dimensions as our human annotation scheme (\S\ref{sec:annotation}). The judge receives the figure image, textual context, and cited evidence. More details on the validation are in \S\ref{sec:llm-judge-validation} and Table~\ref{tab:judge-iaa}. Expert labels select the original SFT items, while the judge supplies labels for the non-expert subset and verifies rephrased variants.

\begin{prompt}[title={Prompt 4: Zero-Shot LLM Judge}]
You are an expert scientist evaluating a question-answer pair grounded in a research paper figure.\\

Your task is to evaluate each question-answer pair grounded in a research paper figure. Focus on the questions' usefulness, clarity, correctness, and whether the answer truly follows from the cited source text. We're looking for questions that are not directly answered in the figures or their captions themselves, but are triggered by the figures, and answered later in the paper.\\

\textbf{Paper Context:} Title: \texttt{\{title\}}. Paper ID: \texttt{\{paper\_id\}}.\\
\textbf{Figure Caption:} \texttt{\{figure\_info\}}\\
\textbf{Cited Source Paragraph(s):} \texttt{\{source\_content\}}\\
\textbf{Question:} \texttt{\{question\}}\\
\textbf{LLM Answer:} \texttt{\{answer\}}\\

Look at the figure image provided. Read the title and think about the key research question(s) in this paper. Look at the figure + caption, and read the question. Read the answer and the cited source paragraph(s). Score each criterion below.\\

\textbf{Dimensions:}
\begin{enumerate}[nosep, leftmargin=*]
  \item \textbf{Question quality} (grammar and clarity): ``perfect'' $|$ ``minor'' $|$ ``major''
  \item \textbf{Question salience}: ``very'' $|$ ``somewhat'' $|$ ``not''
  \item \textbf{Answer quality}: ``4'' (excellent) $|$ ``3'' (good) $|$ ``2'' (fair) $|$ ``1'' (poor)
  \item \textbf{Answer correct?}: ``yes'' $|$ ``partial'' $|$ ``no''
  \item \textbf{Figure usefulness}: ``essential'' $|$ ``helpful'' $|$ ``not''
  \item \textbf{Answered by figure + caption?}: ``yes'' $|$ ``no''
  \item \textbf{Figure type}: ``result'' $|$ ``method'' $|$ ``data'' $|$ ``comparison'' $|$ ``other''
\end{enumerate}
~\\
\textbf{Output:} JSON with \texttt{q-grammar}, \texttt{salience}, \texttt{answer\_quality}, \texttt{answer-correct}, \texttt{figure-useful}, \texttt{answered-by-figure}, \texttt{figure-type}, \texttt{notes}.
\end{prompt}

\section{Question depth analysis}
\label{app:depth-analysis}

To characterize how \mqud{} questions differ from existing benchmarks, we classify 100 random questions per dataset along two independent dimensions using GPT-5-mini.

\paragraph{Cognitive level.} We classify questions into recall (retrieving facts or values), analytical (comparing, finding patterns, computing), and evaluative (judging, assessing implications). Table~\ref{tab:depth-blooms} shows that \mqud{} questions concentrate at the analytical level (76\%), while ChartQA is predominantly recall (63\%). SPIQA occupies an intermediate position.

\begin{table}[h]
\centering
\small
\setlength{\tabcolsep}{4pt}
\begin{tabular}{lccc}
\toprule
\textbf{Level} & \textbf{MQUD} & \textbf{SPIQA} & \textbf{ChartQA} \\
\midrule
Recall & 17\% & 31\% & \textbf{63\%} \\
Analytical & \textbf{76\%} & 48\% & 37\% \\
Evaluative & 7\% & 21\% & 0\% \\
\bottomrule
\end{tabular}
\caption{Cognitive level distribution ($n{=}100$ per dataset, GPT-5-mini). Levels correspond to Anderson \& Krathwohl's (2001) revised taxonomy: recall = remember+understand, analytical = apply+analyze, evaluative = evaluate+create.}
\label{tab:depth-blooms}
\end{table}

\paragraph{Question-word distribution.} As a transparent, classifier-free measure, we bin questions by their leading question word into explanatory or relational forms (why, how-process, consequence, extent, comparison) and extraction-oriented forms (what-is, which, how-many, yes/no). This lexical grouping is separate from the annotation-based labels in \S\ref{sec:annotation}. Explanatory or relational forms account for 76\% of \mqud{} questions, compared with 29\% for SPIQA and 4\% for ChartQA.

\section{Qualitative grounding examples}
\label{app:qualitative-grounding}

We present three examples from the validation set, showing how the figure, the question, and the paper's answer fit together.

\begin{table}[t]
\centering
\small
\begin{tabular}{@{}llp{0.66\linewidth}@{}}
\toprule
\textbf{Type} & \textbf{Model} & \textbf{Question} \\
\midrule
\multirow{2}{*}{Extent} & Base & ``How does CRISP compare to the theoretical ideal unlearning point?'' \\
& SFT & ``How does CRISP's performance cluster near the ideal unlearning point compared to RMU and ELM?'' \\
\midrule
\multirow{2}{*}{Concept} & Base & ``How does the After Intervention image represent the effect of steering vectors?'' \\
& SFT & ``How does the intervention alter the model's focus from text to visual elements?'' \\
\bottomrule
\end{tabular}
\caption{Representative base and SFT generated questions for the same figures.}
\label{tab:qualitative}
\end{table}

\begin{prompt}[title={Example 1: Extent question (answered by the figure)}]
\textbf{Paper:} \emph{Persistent Concept Unlearning via Sparse Autoencoders}\\[4pt]
\begin{center}
\includegraphics[width=0.65\linewidth]{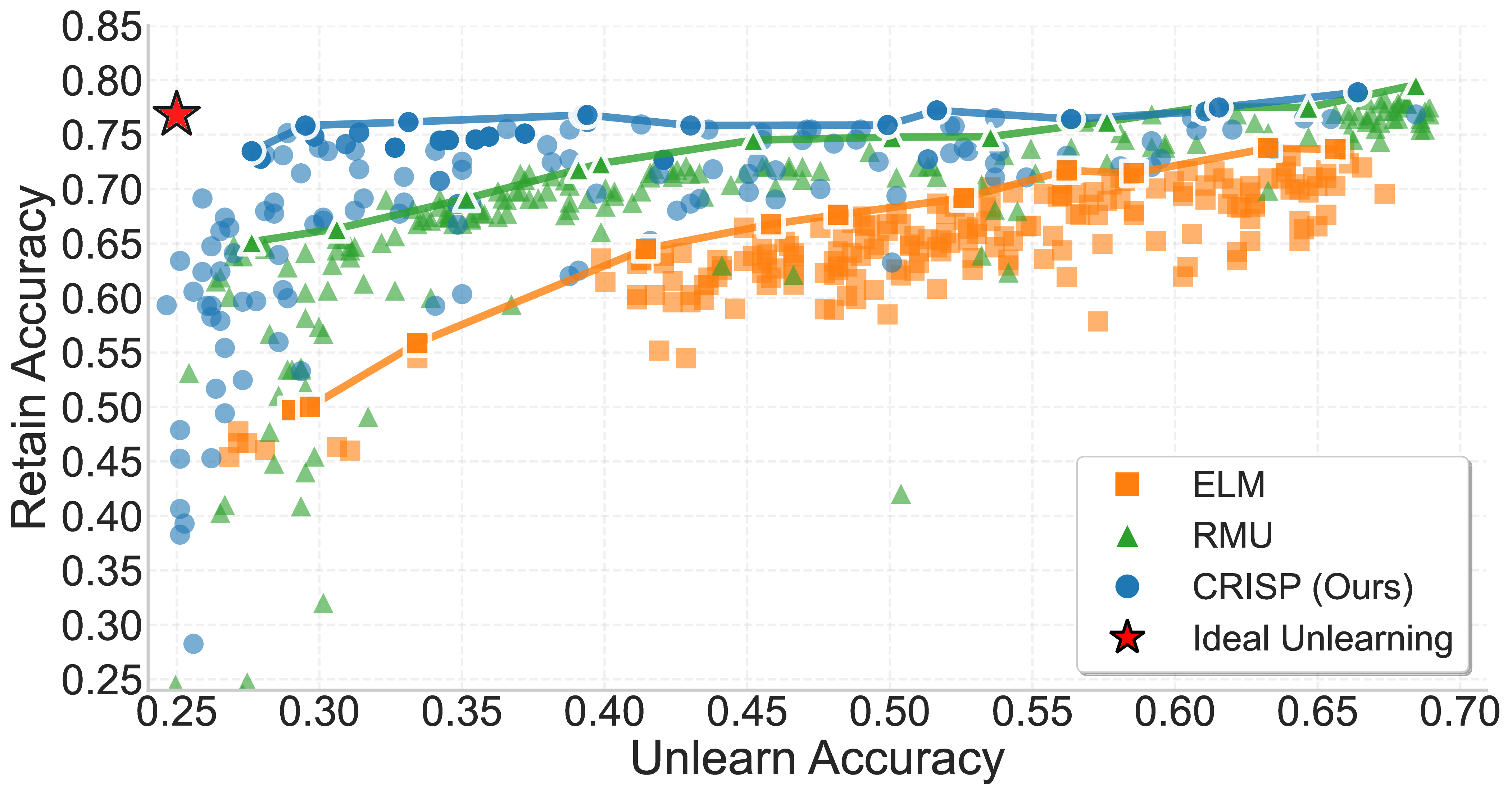}
\end{center}
\textbf{Reference question:} \emph{To what extent does CRISP outperform ELM in retaining accuracy across the datasets shown?}\\
\textbf{Answer:} CRISP vastly outperforms ELM: its points cluster at the high end of the retain-accuracy axis, showing that persistent concept unlearning balances the unlearning trade-off while retaining more knowledge.\\[4pt]
\textbf{Source evidence:} \emph{The paper describes the unlearn--retain trade-off and reports that CRISP achieves Pareto-dominant performance for both criteria.}\\[4pt]
\end{prompt}

\begin{prompt}[title={Example 2: Cause QUD (multimodal QUD)}]
\textbf{Paper:} \emph{IsoScore: Measuring the Uniformity of Embedding Space Utilization}\\[4pt]
\begin{center}
\includegraphics[width=0.65\linewidth]{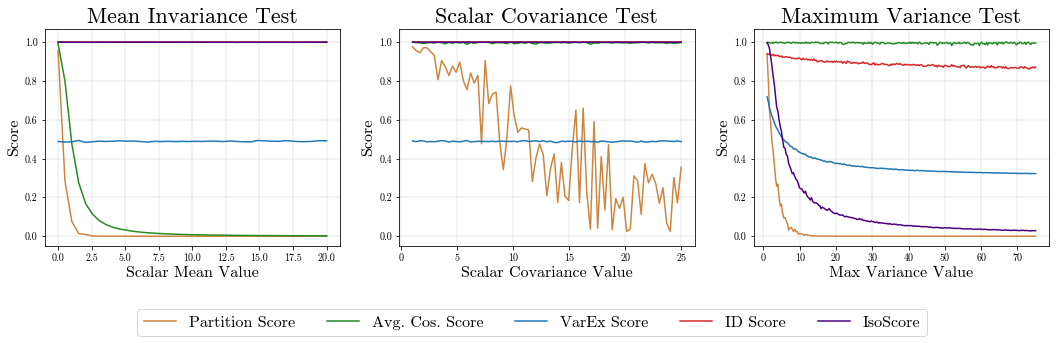}
\end{center}
\textbf{Reference question:} \emph{Why does the Partition Score decrease so sharply in the Mean Invariance Test panel as the scalar mean value increases?}\\
\textbf{Answer:} The Partition Score is not mean-agnostic: non-zero-mean data skews it downward, so it cannot measure isotropy reliably when the mean shifts.\\[4pt]
\textbf{Source evidence:} \emph{The paper's mean-agnosticism test shows the partition score is not mean-agnostic, whereas IsoScore is, being a function of the covariance matrix alone.}\\[4pt]
\end{prompt}

\begin{prompt}[title={Example 3: Cause QUD (multimodal QUD)}]
\textbf{Paper:} \emph{Multilingual Code Co-Evolution Using Large Language Models}\\[4pt]
\begin{center}
\includegraphics[width=0.65\linewidth]{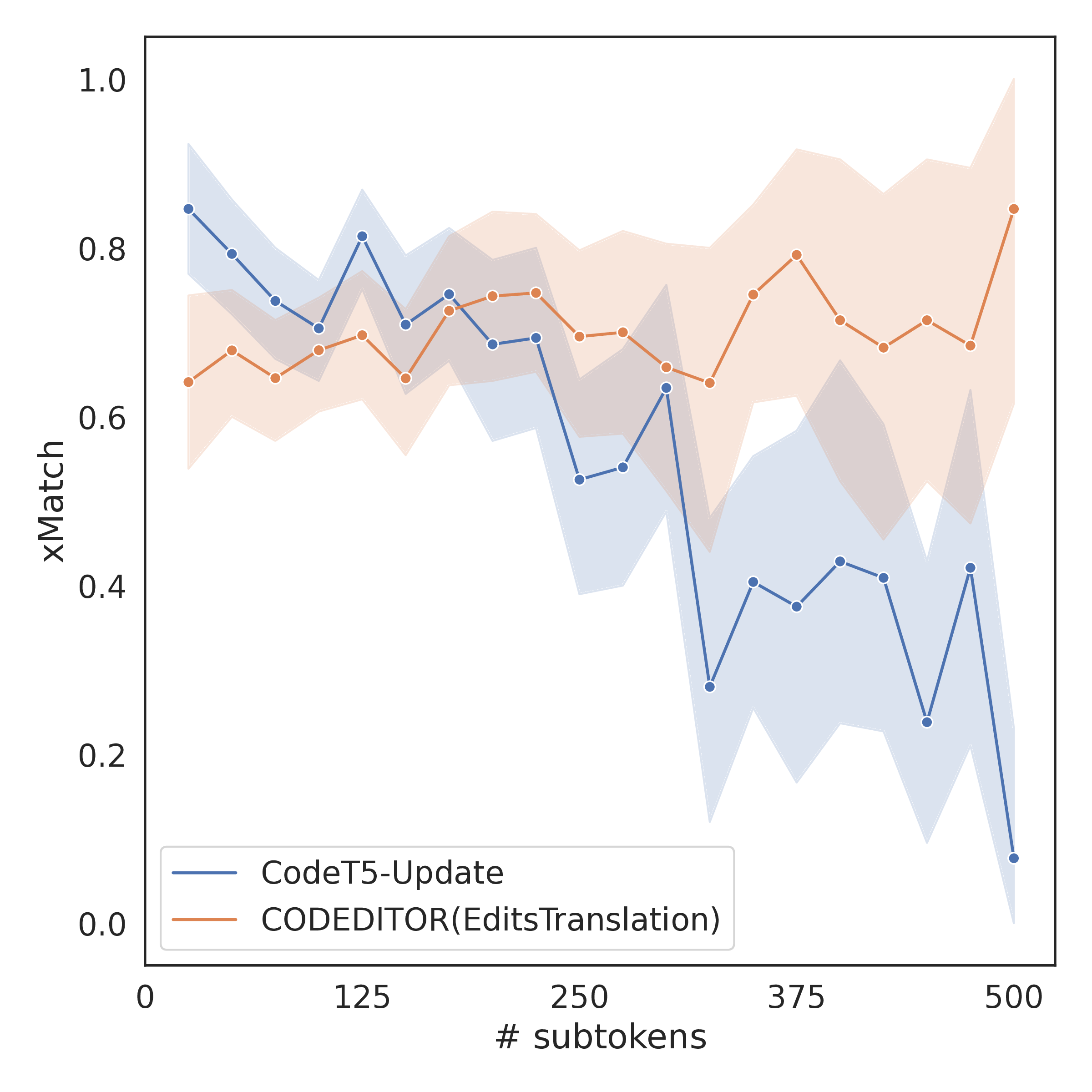}
\end{center}
\textbf{Reference question:} \emph{Why does CodeT5-Update's performance drop significantly as the number of subtokens increases, unlike CODEDITOR?}\\
\textbf{Answer:} CodeT5-Update generates the entire new code, which becomes harder as inputs grow; CODEDITOR transforms edits instead, so its performance stays stable on longer inputs.\\[4pt]
\textbf{Source evidence:} \emph{The paper breaks down accuracy by input length and attributes CODEDITOR's stability to modeling edits rather than generating whole code.}\\[4pt]
\end{prompt}

\section{Qualitative comparison}
\label{app:qualitative}

\begin{figure}[p]
\centering
\includegraphics[width=\linewidth,height=0.9\textheight,keepaspectratio]{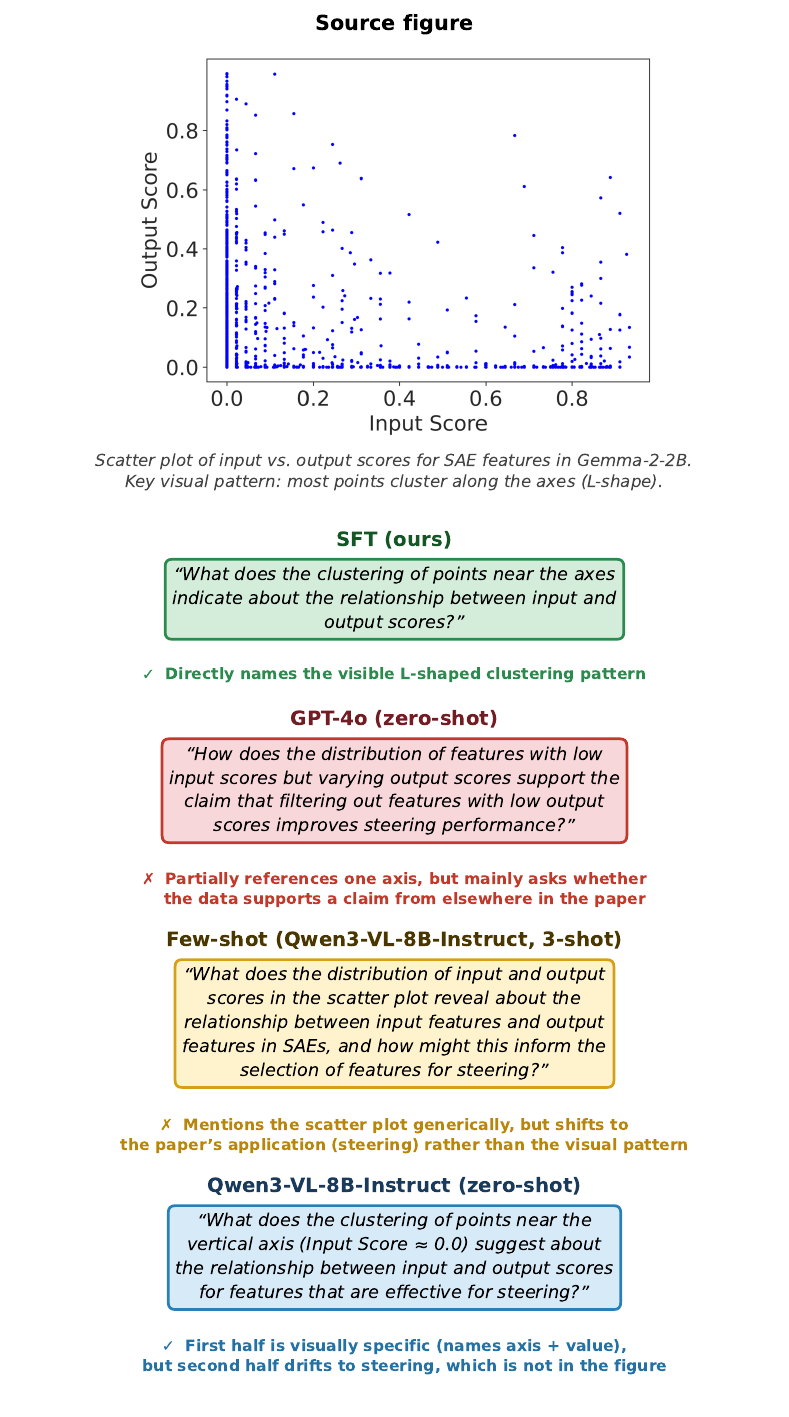}
\caption{Qualitative comparison of generated questions for a scatter plot. \textbf{Top:} source figure showing input vs.\ output scores for SAE features. \textbf{Bottom:} questions from four methods with annotations. SFT captures the full visual pattern; other methods partially ground in the figure but drift toward abstract or paper-level claims.}
\label{fig:qualitative-comparison}
\end{figure}

Figure~\ref{fig:qualitative-comparison} compares questions from four methods for the same figure. Only SFT names the full visible pattern; the other methods drift toward paper-level claims. Aggregate evidence is in \S\ref{sec:grounding-results}.

\section{Training details}
\label{app:training}

We fine-tune Qwen3.5-9B using LoRA on all linear layers with the vision encoder frozen. Training curves are in Figure~\ref{fig:training-curves}. We select the checkpoint with lowest validation loss.

\begin{table}[h]
\centering
\small
\begin{tabular}{lc}
\toprule
\textbf{Hyperparameter} & \textbf{Value} \\
\midrule
LoRA rank & 64 \\
LoRA $\alpha$ & 128 \\
LoRA dropout & 0.05 \\
Learning rate & $1 \times 10^{-4}$ (cosine, 0.1 warmup) \\
Epochs & 2 \\
Effective batch size & 16 \\
Max sequence length & 2048 \\
Hardware & 2$\times$ NVIDIA A40 \\
\bottomrule
\end{tabular}
\caption{SFT hyperparameters. Data splits described in \S\ref{sec:sft}.}
\label{tab:hyperparams}
\end{table}

\clearpage
\end{document}